\documentclass[conference]{IEEEtran}
\usepackage{times}

\usepackage[numbers]{natbib}
\usepackage{multicol}
\usepackage[bookmarks=true]{hyperref}

\usepackage[utf8]{inputenc}
\usepackage{multirow} 
\usepackage[table]{xcolor} 
\usepackage{array}    
\usepackage{bm}         
\usepackage{makecell}   
\definecolor{lightlavender}{RGB}{235, 235, 255}
\usepackage{subfigure}
\usepackage{graphicx} 
\usepackage{amsmath}
\usepackage{amssymb} 
\usepackage{booktabs}

\usepackage{tabularx}

\usepackage{pgfplots}
\pgfplotsset{compat=1.18}
\usepackage{cuted}  
\usepackage{capt-of}

\begin{document}

\title{DexFormer: Cross-Embodied Dexterous Manipulation via History-Conditioned Transformer} 
\author{Ke Zhang$^{1*}$, Lixin Xu$^{1*}$, Chengyi Song$^{1}$, Junzhe Xu$^{1}$, Xiaoyi Lin$^{2}$, Zeyu Jiang$^{1}$, Renjing Xu$^{1\dagger}$
\\
$^{1}$The Hong Kong University of Science and Technology (Guangzhou), 
$^{2}$ Wuhan University
\\
$^{*}$Equal contribution, $^{\dagger}$Corresponding author
}

\maketitle

\begin{abstract}
Dexterous manipulation remains one of the most challenging problems in robotics, requiring coherent control of high-DoF hands and arms under complex, contact-rich dynamics. A major barrier is embodiment variability: different dexterous hands exhibit distinct kinematics and dynamics, forcing prior methods to train separate policies or rely on shared action spaces with per-embodiment decoder heads. We present DexFormer, an end-to-end, dynamics-aware cross-embodiment policy built on a modified transformer backbone that conditions on historical observations. By using temporal context to infer morphology and dynamics on the fly, DexFormer adapts to diverse hand configurations and produces embodiment-appropriate control actions. Trained over a variety of procedurally generated dexterous-hand assets, DexFormer acquires a generalizable manipulation prior and exhibits strong zero-shot transfer to Leap Hand, Allegro Hand, and Rapid Hand. Our results show that a single policy can generalize across heterogeneous hand embodiments, establishing a scalable foundation for cross-embodiment dexterous manipulation. Project website: \href{https://davidlxu.github.io/DexFormer-web/}{https://davidlxu.github.io/DexFormer-web/}.
\end{abstract}

\IEEEpeerreviewmaketitle

\section{Introduction}
\label{sec:introduction}
Dexterous manipulation is a key capability for general-purpose robotic manipulation. Multi-fingered robotic hands enable a wide range of functional behaviors, spanning lifting and reorienting everyday objects~\cite{DextrAH-G,DextrAH-RGB,GraspXL}, grasping in clutter~\cite{DexSinGrasp,ClutterDexGrasp}, non-prehensile manipulation~\cite{nonprehensile,ExDex}, and in-hand reorientation and fine pose adjustment~\cite{DexNDM,Visual_Dexterity,SpinningPen,Robot_Synesthesia}. In practice, however, when transferring to each new hand embodiment typically requires carefully training and tuning a policy from scratch. Different hands induce distinct kinematics, inertias, actuator responses, and contact dynamics, so the same control strategy can lead to drastically different behavior. This embodiment-specific dynamics makes learning and sim-to-real transfer brittle and expensive, often requiring repeated system identification, retuning, or additional real-world data collection for every new platform~\cite{DexNDM,Visual_Dexterity}. 

The sim-to-real gap is already substantial for a single robot, and becomes even more severe when transferring across different dexterous hand embodiments. Broadly, existing approaches to bridge this gap fall into three paradigms. The first is system identification (SysID), which fits simulator parameters to real hardware through data-driven model identification or active exploration, reducing parameter mismatch between simulation and reality~\cite{lee2023robot,sobanbabu2025sampling}. While effective, SysID is limited by the chosen model parameterization and typically needs to be repeated for each new embodiment.
The second paradigm is domain randomization, which trains policies over wide distributions of physical and morphological parameters so that the real world appears as just another sample from training. Massive randomization enabled sim-to-real dexterous in-hand manipulation without real data collection~\cite{andrychowicz2020learning}, and large-scale RL over procedurally generated robots has produced generalist locomotion policies that control previously unseen robots zero-shot without explicit kinematic knowledge~\cite{liu2025locoformer}. However, randomization alone does not provide an explicit mechanism to compensate for embodiment-specific dynamics at test time.
The third paradigm learns residual or compensatory models for unmodeled dynamics on top of simulation. Methods such as~\cite{fey2025bridging} learns unsupervised actuator networks to capture nonlinear hardware effects. Similarly, DexNDM~\cite{DexNDM} learns a joint-wise neural dynamics model from real interaction data and trains a residual controller on top of a simulation policy, enabling robust dexterous manipulation without precise object state estimation. These residual approaches can significantly narrow the sim-to-real gap, but depend on collecting embodiment-specific real-world data and therefore do not directly address zero-shot cross-embodiment transfer.

\begin{figure}[t!]
    \centering
    \includegraphics[width=\linewidth]{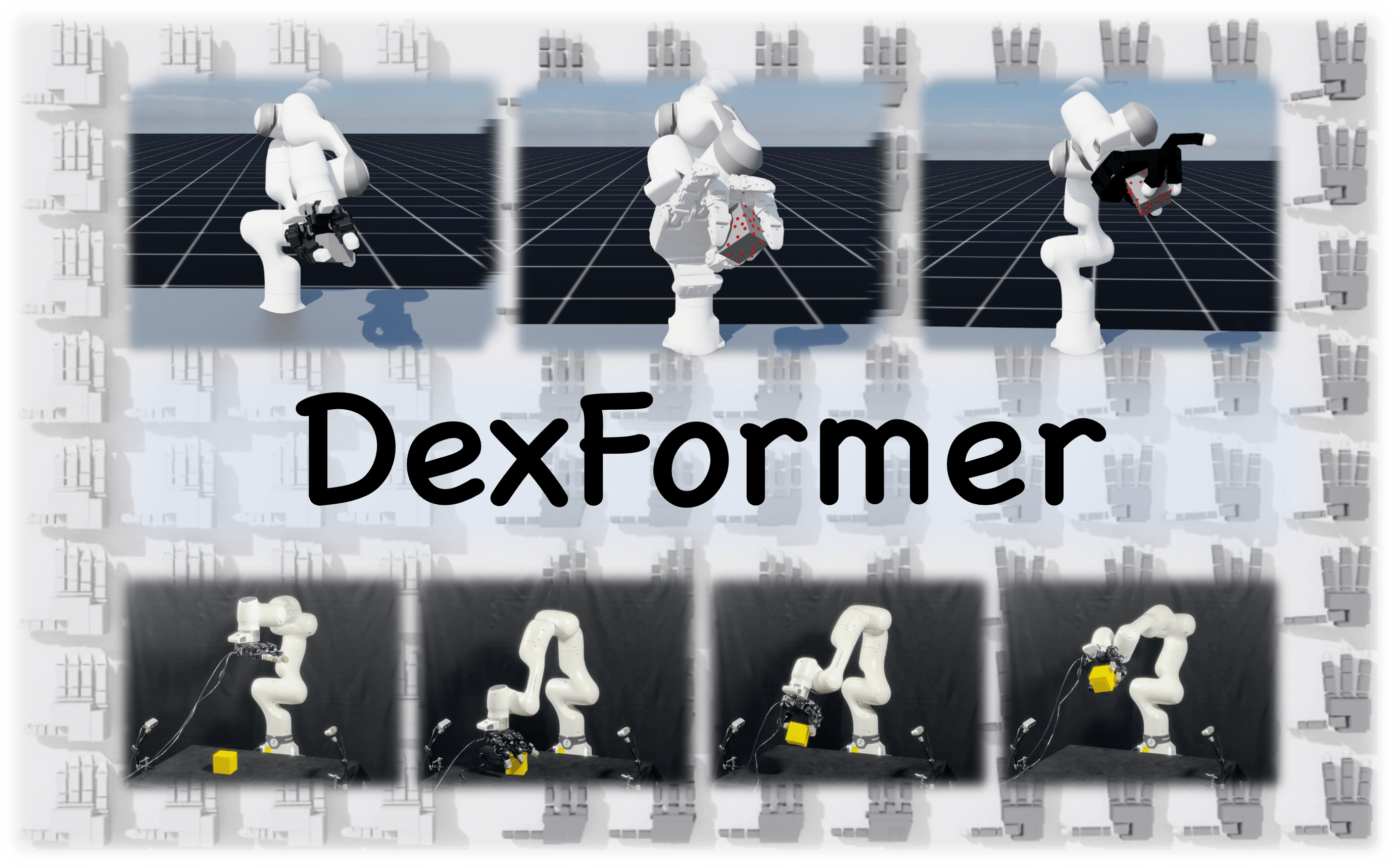}
    \caption{DexFormer learns a unified history-conditioned policy that transfers dexterous grasping across diverse hand embodiments, enabling zero-shot deployment from large-scale simulation to real-world robots.}
    \label{fig:placeholder}
\end{figure}

In this work, we introduce \textbf{DexFormer}, a cross-embodiment dexterous manipulation policy based on a history-conditioned transformer. DexFormer leverages temporal context to perform implicit morphology inference from observation histories, enabling the policy to adapt online to different hand dynamics without explicit morphology identifiers or embodiment-specific heads. We construct a broad morphology distribution via procedural randomization of canonical dexterous hands and demonstrate that a single morphology-agnostic policy trained on randomized embodiments can \emph{zero-shot} generalize to unseen canonical hands and their variants. DexFormer thus represents a scalable approach to cross-embodiment dexterity and reveals a promising connection between large-scale RL training and history-conditioned models capable of learning latent physical structure.

Our main contributions are summarized as follows:
\begin{itemize}
    \item We introduce DexFormer, a history-conditioned transformer trained on large scale of hand embodiments, which performs implicit morphology inference for cross-embodiment adaptive control;
    \item We develop a large-scale morphology-randomization pipeline and a distributed training framework for dexterous hands cross embodiment training;
    \item We show that a single morphology-agnostic policy can generalize to unseen dexterous embodiments and multiple real dexterous hands without manual retargeting, explicit morphology encodings, or separate policy heads.
\end{itemize}

\section{Related Work}
\label{sec:related_works}
\subsection{Cross-embodiment dexterous manipulation}

Cross-embodiment dexterous manipulation has been approached from two main directions. One line of work generates static cross-hand grasp poses followed by open-loop execution, using unified contact representations and optimization or physics solvers to produce feasible grasps across morphologies \cite{wei2024d,CEDex,zhang2025cross}; however, such open-loop strategies struggle to react to disturbances and real-time contact changes during execution. Others learn embodiment-aware closed-loop controllers, either by explicitly encoding hand kinematic graphs and distilling experts into a single zero-shot policy \cite{patel2025get}, or by defining an eigen-grasp action space that transfers policies to new hands via retargeting mapping \cite{yuan2024cross}; these approaches depend on explicit kinematic modeling or retargeting mapping and are difficult to extend across fundamentally different canonical hand types. We instead learn a history-conditioned policy in a canonical shared action space that implicitly infers embodiment-specific dynamics from temporal context, enabling closed-loop, zero-shot transfer across heterogeneous hands without explicit morphology encoding or retargeting.

\subsection{Dynamics-aware manipulation}
Methods in this direction explicitly model or compensate for mismatches between simulated and real dynamics. Unsupervised Actuator Networks (UAN)~\cite{fey2025bridging} learn data-driven corrections to simulator actuator models from real-world trajectories, improving sim-to-real transfer by reducing exploitation of simulator inaccuracies in highly dynamic loco-manipulation. DexNDM~\cite{DexNDM} instead learns a joint-wise neural dynamics model from autonomously collected real interaction data and uses it to train a residual controller on top of a simulation policy, enabling robust in-hand rotation across diverse objects without precise object state estimation. Complementarily, Zhao et al. \cite{2601.02778} narrows the sim-to-real gap by enriching simulation with calibrated actuator and tactile dynamics, combining high-fidelity tactile simulation, current–torque calibration, and randomized motor nonidealities to train force-aware policies that deploy zero-shot on real hardware. These approaches rely on either explicit real-world dynamics learning or highly accurate simulator modeling. Our method avoids additional dynamics models or residual adaptation by training a single policy over a broad, randomized morphology and dynamics distribution and letting temporal context implicitly capture embodiment-specific behavior.

\subsection{Test-time adaptation} 
Test-time and online adaptation has been extensively studied in legged locomotion and is beginning to appear in manipulation. Rapid Motor Adaptation (RMA) \cite{kumar2021rma} learns a latent embedding online from recent state–action history and conditions a base policy on this latent embedding to rapidly compensate for unmodeled dynamics without additional real-world rollouts or calibration. Variants of this idea have been applied to bipedal robots~\cite{9981091}, manipulator arms~\cite{Liang_2024_CVPR}, and dexterous in-hand rotation~\cite{qi2023hand}, showing that short temporal histories can support fast system identification and closed-loop correction under changing contacts and loads. In parallel, in-context adaptation with sequence models such as LocoFormer \cite{liu2025locoformer} conditions policies directly on history to achieve zero-shot adaptation across terrains and tasks without an explicit identification module. In our setting, we do not explicitly learn a separate adaptation module to suit each environment, but instead learn one history-conditioned policy to a wide distribution of procedurally randomized hand morphologies during training, so that it implicitly infers embodiment-specific dynamics from temporal context and can zero-shot generalize to unseen hand embodiments.

\section{Method}
\label{sec:methods}
\subsection{Problem Formulation}
We model dexterous manipulation as a POMDP in which a robotic hand must grasp and stably manipulate objects drawn from a task distribution. Each episode samples an object $o \in \mathcal{O}$ and a hand embodiment $e \in \mathcal{E}$, inducing morphology-dependent transition dynamics. 

We aim to learn a morphology-agnostic policy $\pi_\theta$ that receives observations but does not observe the embodiment identity. Let $o_t$ denote the observation of the system at time $t$, including the hand, arm, and object pointcloud, and $a_t$ denote the action selected by the policy at time $t$. The reinforcement learning objective is to maximize the discounted reward over the joint distribution of embodiments and objects:
\begin{equation}
\max_{\theta} \;
\mathbb{E}_{e \sim p(\mathcal{E}),\, o \sim p(\mathcal{O}),\, \tau \sim \pi_\theta}
\left[ \sum_{t=0}^{T} \gamma^t R(o_t, a_t) \right],
\label{eq:rl_objective}
\end{equation}
where $R$ rewards successful grasping behaviors. Since the embodiment is unobserved, the agent must infer morphology-dependent dynamics through temporal interaction and adapt its control accordingly.

\subsection{Shared Action Space}

\begin{figure*}[h!]
    \centering
    \subfigure[Shared action space and anatomical correspondence.]{
        \includegraphics[width=0.39\linewidth]{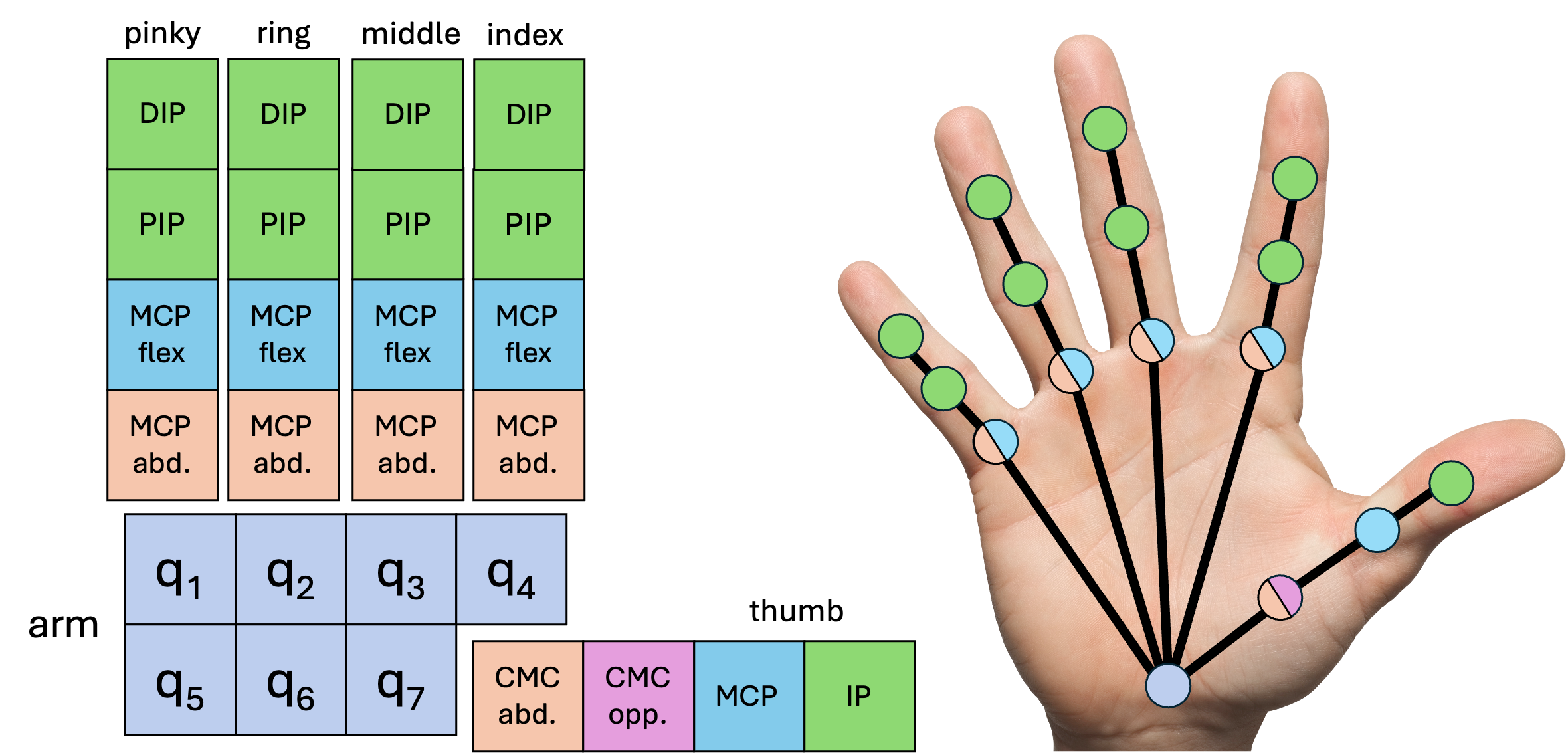}
    }
    \vspace{4pt}
    \subfigure[Shared action space with masks for specific embodiments.]{
        \includegraphics[width=0.57\linewidth]{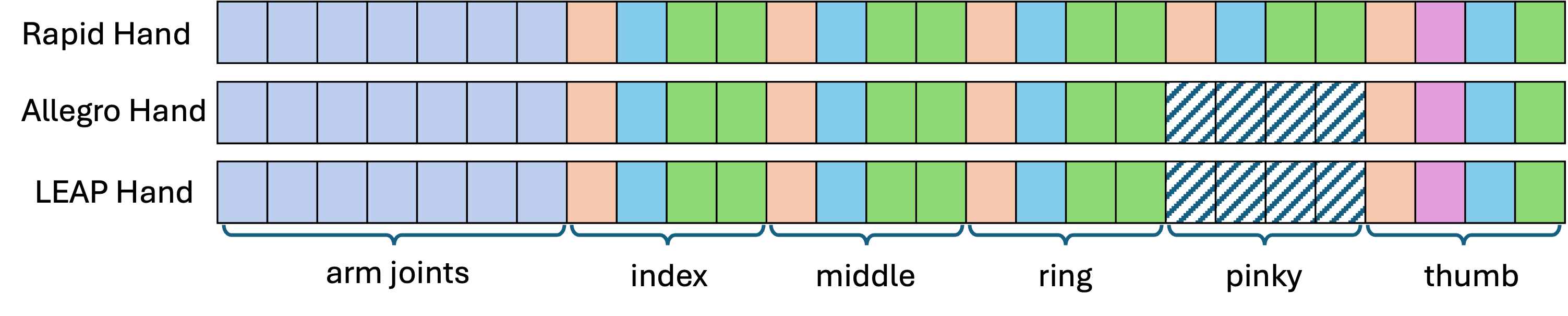}
    }
    \caption{Shared action space. (a) The canonical action space (left) defines a morphology-invariant embedding in which joints with the same anatomical function share fixed indices in the action space. MCP (blue/orange) governs flexion and abduction, PIP/DIP (green) provide proximal/distal flexion, while the thumb uses a distinct structure where CMC (orange/purple) supports abduction and opposition, MCP (blue) controls basal flexion, and IP (green) provides distal flexion. (b) Canonical embedding flattened into 20-Dim space. Lower-DoF hands such as LEAP and Allegro hands zero-pad unused canonical dimensions, where as higher-DoF embodiments like Rapid Hand fully populate the embedding, enabling shared control across heterogeneous embodiments.}
    \label{fig:action_space}
\end{figure*}

We consider a family of dexterous robotic hands with different numbers of actuated finger joints. Let $\mathcal{E}$ denote the set of hand embodiments and, for each embodiment $e \in \mathcal{E}$, let $d_e$ be the number of actuated finger joint DoF. To enable a unified control interface across heterogeneous hands, we define a canonical $D_F$-dimensional finger action space, where joints with the same functional role (e.g., thumb abduction, index flexion) are assigned to the same canonical indices across embodiments.

For embodiment $e$ with $d_e$ actuated finger joints, we denote its native finger command by $\tilde{a}_t^F(e) \in \mathbb{R}^{d_e}$. We embed this command into the canonical action space via a fixed embedding operator
\[
P_e : \mathbb{R}^{d_e} \rightarrow \mathbb{R}^{D_F},
\]
which writes the $d_e$ joint commands into embodiment-specific canonical indices corresponding to their functional roles, such as MCP abduction and flexion ordering, and sets all remaining entries to zero:
\[
a_t^F = P_e\big(\tilde{a}_t^F(e)\big) \in \mathbb{R}^{D_F}.
\]
Embodiments with fewer finger DoF use zero-padding for unused canonical dimensions, while higher-DoF hands may occupy all or most of the $D_F$ indices, as shown in Fig.~\ref{fig:action_space}.

We further apply temporal smoothing to obtain the executed finger action. Let $\hat{a}_t^F \in \mathbb{R}^{D_F}$ denote the raw finger action output by the policy. The smoothed finger command is defined as
\begin{equation}
a_t^F = \lambda\, \hat{a}_t^F + (1 - \lambda)\, a_{t-1}^F,
\label{eq:finger_smoothing}
\end{equation}
where $\lambda \in (0,1]$ controls the degree of action smoothing.

The overall high-level action at time $t$ is then given by
\begin{equation}
a_t \triangleq \left[ a_t^F,\, a_t^A \right] \in \mathbb{R}^{D_F + D_A},
\label{eq:shared_action}
\end{equation}
where $a_t^A \in \mathbb{R}^{D_A}$ is the robotic arm delta pose (e.g., $D_A = 7$ for a 7-DoF arm). This shared high-level action is consumed by embodiment-specific low-level controllers, which map $a_t$ to torques or joint targets; zero-padded finger dimensions are simply ignored for lower-DoF hands. In our experiments (Section~\ref{sec:experiments}), we instantiate this formulation with a specific choice of $D_F$ and a concrete set of hand embodiments.

\subsection{Embodiment Generation}
We construct a family of dexterous hand embodiments by perturbing morphology-related physical parameters of canonical hands. Let $\mathcal{E}_{\mathrm{canon}}$ denote the set of canonical embodiments and, for each $e \in \mathcal{E}_{\mathrm{canon}}$, let $z(e)$ represent its morphology parameters including link lengths, masses, and inertias. We sample randomized embodiments according to
\[
z^{(k)}(e) \sim p_{\mathrm{morph}}(z \mid e),
\quad k = 1, \dots, N_e,
\]
where $p_{\mathrm{morph}}$ is a perturbation distribution defined over morphology parameters. Each sampled morphology $z^{(k)}(e)$ induces a new embodiment $e^{(k)}$, while preserving the original kinematic graph topology and actuation structure. 

The union of these embodiments
\[
\mathcal{E}_{\mathrm{train}} = \bigcup_{e \in \mathcal{E}_{\mathrm{canon}}} \{e^{(1)}, \dots, e^{(N_e)}\}
\]
forms the morphology-rich training set used for reinforcement learning.

Training on $\mathcal{E}_{\mathrm{train}}$ exposes the policy to a continuum of physical realizations, promoting robustness and improving zero-shot transfer to unseen canonical embodiments at evaluation time.

\subsection{History-Conditioned Transformer}
To infer latent morphology parameters that are unobservable from a single state, we condition the policy on a finite history window. Let $h_t$ denote the history available at time $t$, consisting of past observation-action pairs and the current observation:
\[
h_t = \{ o_k \}_{k=t-H+1}^{t} 
\]
We process this history as a sequence of $H$ temporal tokens. For each timestep $k \in [t-H+1, t]$, we construct tokens by using the embeddings of the observation $o_k$.

\begin{figure}[h!]
    \centering
    \includegraphics[width=0.98\linewidth]{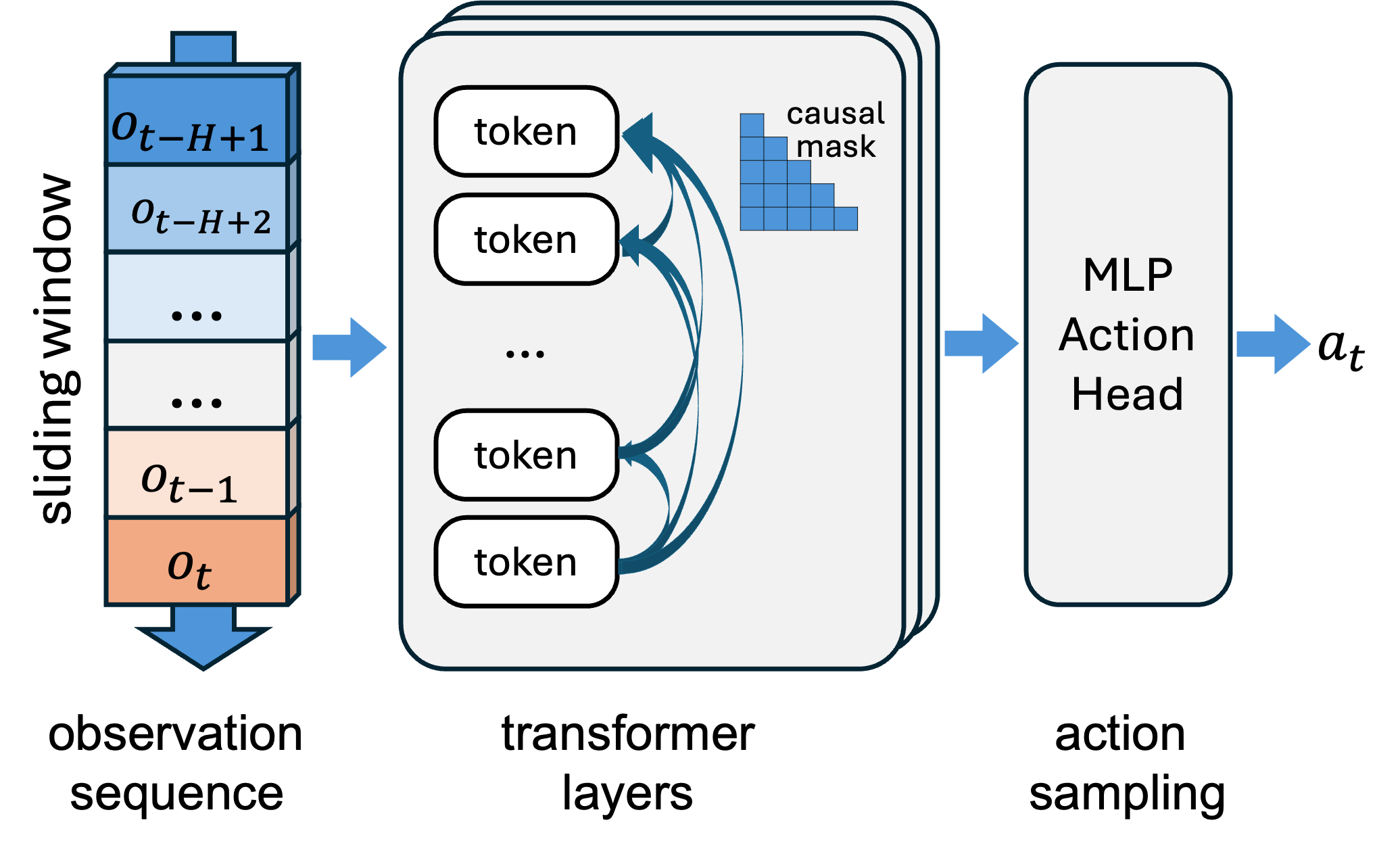}
    \caption{History-conditioned transformer policy architecture. At each timestep, a fixed-length history of observations with horizon $H$ is provided as input and tokenized into a sequence of $H$ tokens. The token sequence is processed by a stack of three transformer layers with positional encoding and causal self-attention. The representation of the final embedding, which attends to all preceding history, is extracted and passed to an MLP action head to parameterize a stochastic policy. Actions are then sampled from the resulting distribution for execution.}
    \label{fig:transformer}
\end{figure}

This sequence, supplemented with learned positional encodings, is fed into a Transformer encoder. Crucially, we apply a \textit{causal mask} to the self-attention mechanism, ensuring that the computation for any token $k$ can only attend to preceding tokens $j \le k$. This prevents information leakage from the future. The Transformer produces a sequence of contextualized latent embeddings. The final embedding, corresponding to the current time $t$, serves as a compact summary of the history and is passed through an MLP action head to produce the action distribution for $a_t$, as illustrated in Fig.~ \ref{fig:transformer}.

\subsection{Observation Space}
The observation space of this policy is defined as 

\begin{equation}
    o_t \triangleq  \left[ o_t^{OC}, o_t^{OT}, a_{t-1}, o_t^{JP}, o_t^{JV}, o_t^{FT}, o_t^{FC}, o_t^{V} \right] \in \mathbb{R}^{377},
\end{equation}
where the current object quaternion $o_t^{OC} \in \mathbb{R}^{4}$, the target object pose $o_t^{OT} \in \mathbb{R}^{7}$, the action $a_{t-1}\in \mathbb{R}^{27}$ at the previous time step,  $o_t^{JP} \in \mathbb{R}^{27}$ include the joint positions, $o_t^{JV} \in \mathbb{R}^{27}$ velocities, fingertip and palm states $o_t^{FT} \in \mathbb{R}^{78}$, and contact forces $o_t^{FC}\in \mathbb{R}^{15}$, and object pointcloud $o_t^{V} \in \mathbb{R}^{192}$.

\subsection{Reward Design}
The total reward is composed of five terms:
\begin{equation} \label{eq:reward}
r_t = r_t^{P} + r_t^{D} + r_t^{M} + r_t^{T} + r_t^{S},
\end{equation}
where $r_t^{P}$ penalizes excessive actions to ensure smooth control, 
$r_t^{D}$ encourages the end-effector to approach the object, 
$r_t^{M}$ encourages the fingers to grasp the object, 
$r_t^{T}$ guides pose tracking under contact, 
and $r_t^{S}$ guides pose tracking without contact.

\begin{table}[h!]
    \centering
    \renewcommand{\arraystretch}{1.4}
    \caption{Reward terms and formulations.}
    \vspace{-2mm}
    \label{tab:reward}
    \begin{tabular}{@{}c l@{}}
    \toprule
    Term & \multicolumn{1}{c@{}}{Equation} \\
    \midrule
    $r^{P}$ & $ -\lambda_1 \!\sum_i a_i^2 - \lambda_2 \!\sum_i (a_i - a_i^{\text{prev}})^2$\\[2pt]
    $r^{D}$ & $ \lambda_3(1 - \tanh\!\left(\frac{\max \| \mathbf{p}_{\text{ee}} - \mathbf{p}_{\text{obj}} \|_2}{\sigma}\right))$ \\[2pt]
    $\mathbb{I}_m$ & $\mathbb{I}\!\left[(\|\mathbf{f}_\text{thumb}\|>\tau \lor
     \|\mathbf{f}_\text{index}\|>\tau \lor 
     \|\mathbf{f}_\text{middle}\|>\tau \lor 
     \|\mathbf{f}_\text{ring}\|>\tau)\right]$ \\[2pt]
    $\mathbb{I}_c$ & $\mathbb{I}\!\left[(\|\mathbf{f}_\text{thumb}\|>\tau) \land 
    (\|\mathbf{f}_\text{index}\|>\tau \lor 
     \|\mathbf{f}_\text{middle}\|>\tau \lor 
     \|\mathbf{f}_\text{ring}\|>\tau)\right]$ \\[2pt]
    $r^{M}$ & $\lambda_4\mathbb{I}_m  + \lambda_5\mathbb{I}_c$ \\[2pt]
    $r^{T}$ & $\lambda_6 (1-\tanh\tfrac{\|\mathbf{p}_{\text{obj}}-\mathbf{p}_{\text{des}}\|_2}{\sigma_{p0}})
    \mathbb{I}_c$ \\[2pt]
    $r^{S}$ & $\lambda_7 (1-\tanh\tfrac{\|\mathbf{p}_{\text{des}}-\mathbf{p}_{\text{obj}}\|_2}{\sigma_{p1}})^2   $ \\
    \bottomrule
    \end{tabular}
    \vspace{-3mm}
\end{table}

\noindent
Here, $a_i$ and $a_i^{\text{prev}}$ denote current and previous actions,
$\mathbf{p}_{\text{ee}}$ and $\mathbf{p}_{\text{obj}}$ are the end-effector and object positions, 
$\mathbb{I}_c$ indicates valid contact,
$\mathbb{I}_m$ encourage exposure to object,
$\mathbf{f}_\text{thumb}$, $\mathbf{f}_\text{index}$, $\mathbf{f}_\text{middle}$, and $\mathbf{f}_\text{ring}$ are contact force vectors measured at the thumb, index, middle, and ring fingertips respectively, $\tau=1$ is the force threshold,
and $d_q(\cdot)$ measures quaternion distance.
$\lambda_1{=}0.005$,
$\lambda_2{=}0.005$, 
$\lambda_3{=}2$, 
$\lambda_4{=}0.8$, 
$\lambda_5{=}2$, $\lambda_6{=}14.0$, $\lambda_7{=}20$,
$\sigma_{p0}{=}0.2$ and $\sigma_{p1}{=}0.1$ .

\subsection{Parallelism for Cross-Embodiment Training}
\begin{figure}[h!]
    \centering
    \includegraphics[width=0.98\linewidth]{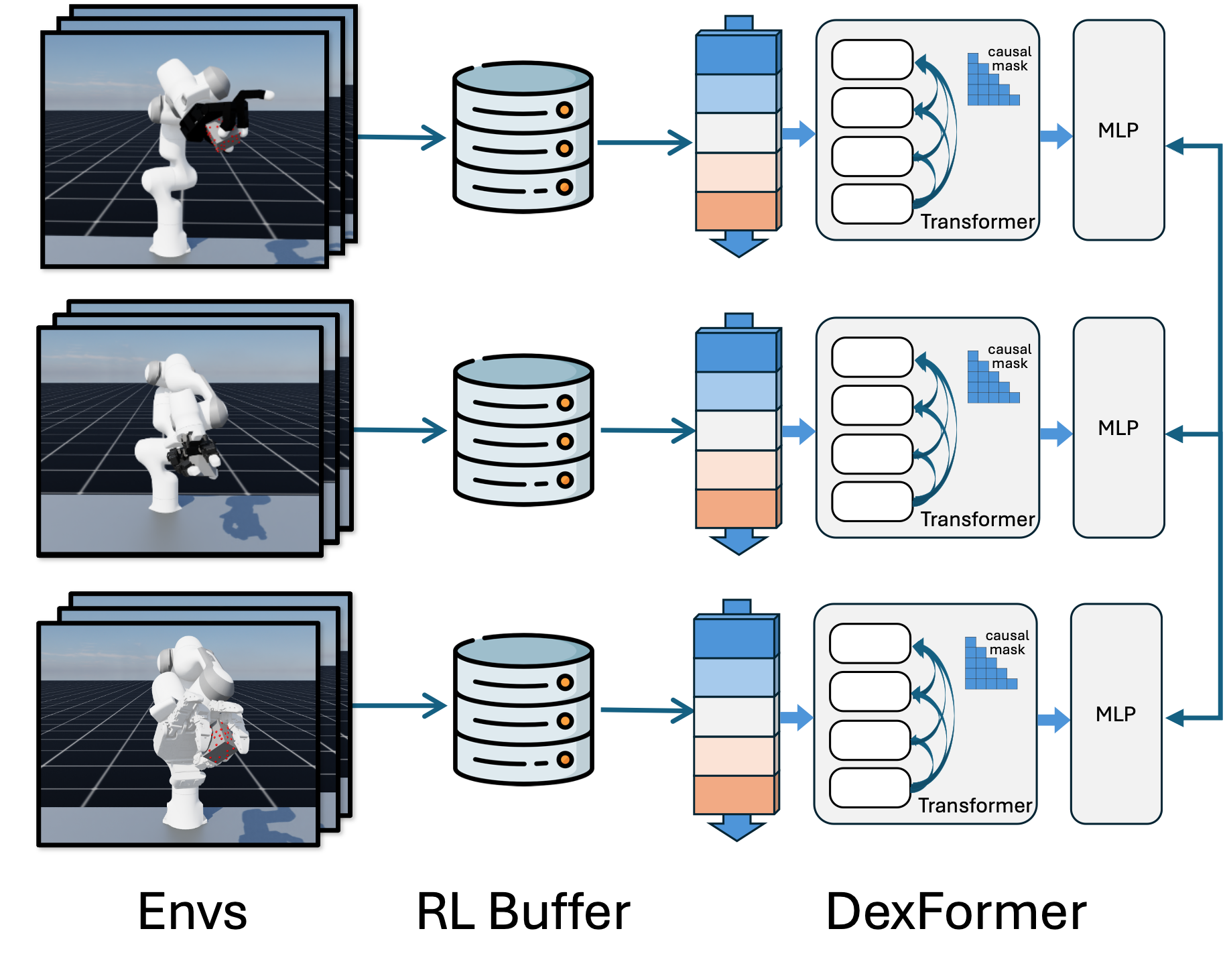}
    \caption{Gradient aggregation during distributed training. While rollouts and forward passes are computed locally on independent GPUs, the all-reduce primitive aggregates gradients during backpropagation, averaging parameter updates to ensure the DexFormer weights remain identical across all devices.}
    \label{fig:all_reduce}
\end{figure}

Simulation platforms such as IsaacLab support parallel environments on a single GPU, it expects similar kinematic topology and actuator structure within each GPU, which is not satisfied across different hand embodiments. To accommodate this, we group randomized variants of the same canonical embodiment on a single GPU and assign morphologically distinct canonical embodiments to different GPUs. Rollouts are performed independently on each GPU, and we employ Distributed Data Parallel (DDP) to maintain a single shared policy: during the trajectory collection and forward pass, each GPU operates locally, while during backpropagation gradients are synchronized all-reduce to ensure consistent parameter updates. This configuration parallelizes morphology diversity across GPUs and environment parallelism within GPUs, enabling scalable cross-embodiment training of a single DexFormer policy, as shown in Fig.~\ref{fig:all_reduce}.

\section{Experiments}
\label{sec:experiments}

This section presents comprehensive simulation and real-world evaluations of DexFormer's performance on unseen embodiments. Our experiments are designed to systematically answer the following research questions: (1) How does DexFormer’s grasp success compare to GRU/LSTM baselines? (2) How well does DexFormer zero-shot generalize across heterogeneous hands? (3) Does historical context improve policy performance? (4) How does embodiment diversity, measured by number of embodiments in training set, affect zero-shot performance? (5) Does performance scale with training parallelism (as measured by number of environment)? (6) How robust is DexFormer to morphological impairments induced by locking individual joints?

\subsection{Implementation Details}
We utilize Isaac Lab~\cite{isaaclab2025} for both training and evaluation. We use Allegro, LEAP \cite{shaw2023leaphand}, and RAPID \cite{wan2025rapid} hands mounted on Franka Arms as the canonical embodiments:
\begin{itemize}
    \item {Franka-Allegro-Canonical}
    \item {Franka-Leap-Canonical}
    \item {Franka-Rapid-Canonical}
\end{itemize}
And then we instantiate morphology-randomized variants of the canonical hands:
\begin{itemize}
    \item {Franka-Allegro-Variants}
    \item {Franka-Leap-Variants}
    \item {Franka-Rapid-Variants}
\end{itemize}
where link lengths and inertial parameters are procedurally perturbed while preserving kinematic and actuation structure.
We leverage a ray-casting pipeline built upon Project Instinct \cite{zhu2026hiking} to generate point-cloud observations of both dexterous embodiments and manipulated objects.

\begin{figure}[h!]
    \centering
    \subfigure[Training embodiment variants (out of 100 for each canonical hand)]{
        \includegraphics[width=0.98\linewidth]{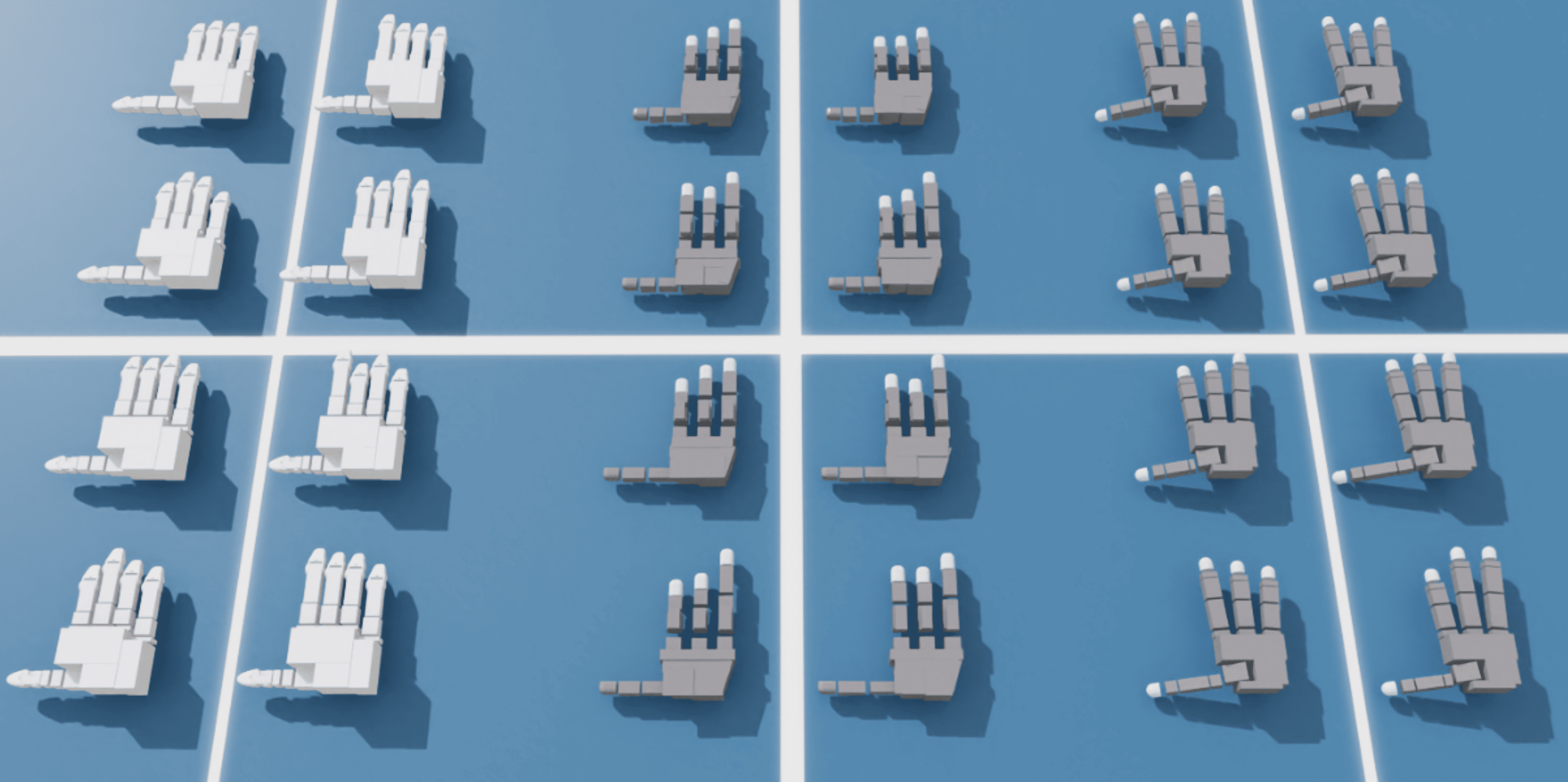}
    \label{fig:training}
    }
    \vspace{4pt}
    \subfigure[Testing embodiment variants (out of 32 for each canonical hand)]{
        \includegraphics[width=0.98\linewidth]{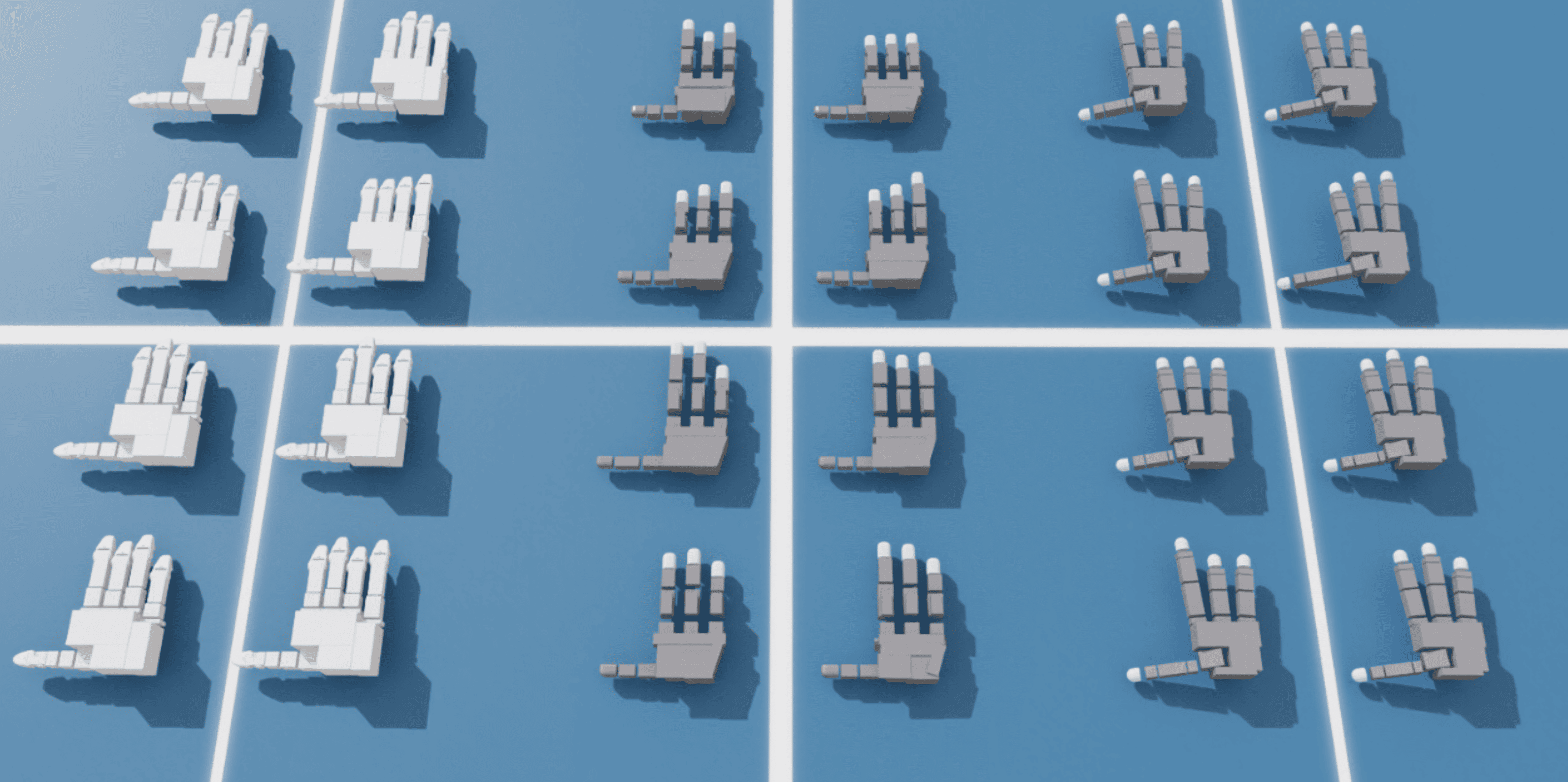}
    \label{fig:testing}
    }
    \caption{Diversified embodiment generation. We synthesize 100 variants per canonical LEAP, Allegro, and RAPID hand for training, and 32 variants per canonical hand for testing. Canonical hands are held out during training and evaluated zero-shot.}
\end{figure}

\begin{figure}[h!]
    \centering
    \includegraphics[width=0.98\linewidth]{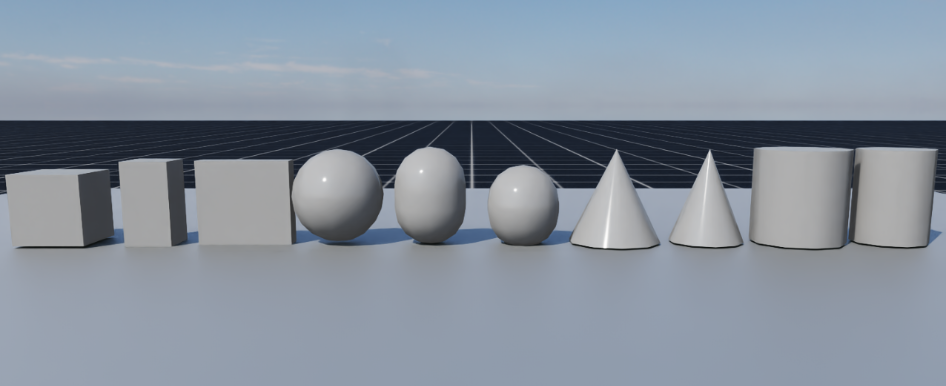}
    \caption{Objects used during training}
    \label{fig:object1}
\end{figure}

We train on 300 randomized embodiments in total, corresponding to 100 variants per canonical hand, shown in Fig.~\ref{fig:training}.  variants form a morphology-rich training distribution for learning a single cross-embodiment policy. Evaluation is conducted in two settings: (i) zero-shot transfer to the three canonical embodiments, and (ii) generalization to $32$ unseen embodiments sampled from the same morphology distribution, shown in Fig.~\ref{fig:testing}.
For each parallel simulation environment, we randomly sample one object from a predefined set of ten objects, shown as in Fig.~\ref{fig:object1}.

Training is performed on three RTX~5090 GPUs. Variants of the Allegro, LEAP, and RAPID hands are assigned to separate GPUs to satisfy batching constraints in Isaac Lab, while Distributed Data Parallel (DDP) synchronizes gradients across GPUs via all-reduce to maintain a single shared policy. We additionally employ Automatic Domain Randomization (ADR)~\cite{akkaya2019solving} with a per-environment scheduler that adjusts difficulty levels (0--10) based on task performance.

\subsection{Metrics and Baselines}
We evaluate policy performance based on \emph{success rate}, defined as the proportion of trials in which the target object reaches the commanded goal position, i.e., $\lVert p^{\text{goal}} - p^{\text{target}} \rVert_2 < 0.05$.

We compare DexFormer against two recurrent policy baselines that use explicit hidden states to encode temporal context:

\begin{itemize}
\item \textbf{LSTM baseline:} a 3-layer LSTM with hidden size 128 is used as the temporal memory module, maintaining cell and hidden states across timesteps to capture longer-range dependencies in the observation history.
\item \textbf{GRU baseline:} a 3-layer GRU with hidden size 128 replaces the LSTM with a lighter gated recurrent architecture, using a single hidden state to model temporal dependencies.
\end{itemize}

\subsection{Main Results.}

\begin{figure*}[ht]
    \centering
    \includegraphics[width=0.32\linewidth]{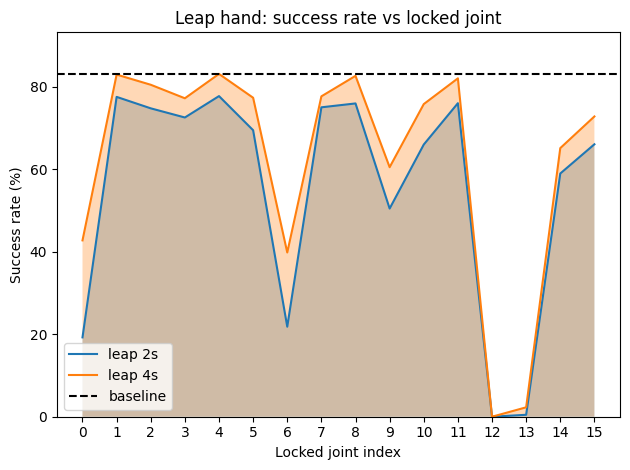}
    \includegraphics[width=0.32\linewidth]{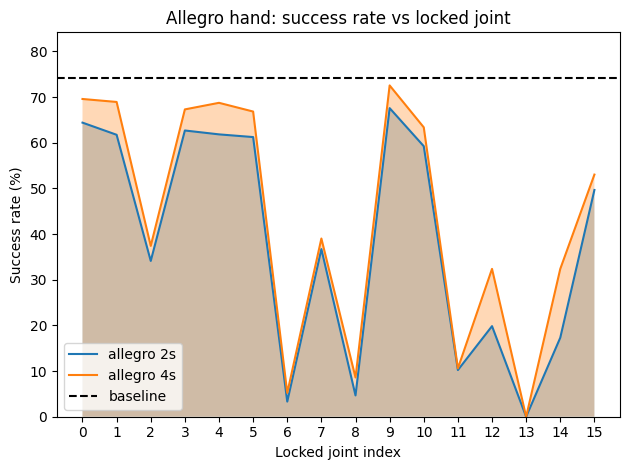}
    \includegraphics[width=0.32\linewidth]{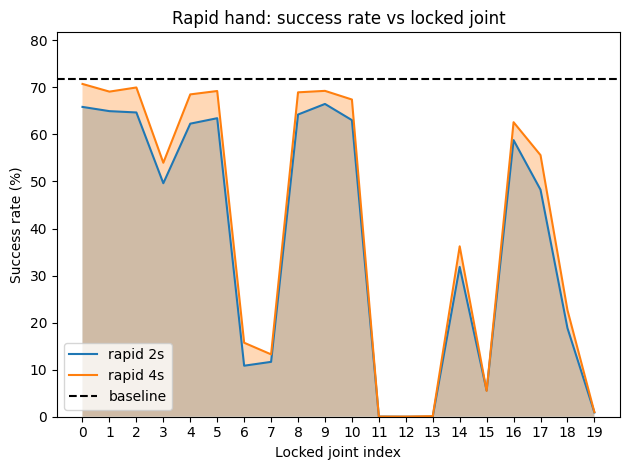}
    \caption{Zero-shot performance under joint locking. For LEAP and Allegro, joint indices 0–15 are sequentially locked, and for RAPID 0–19. We report success rates for 2-s and 4-s execution windows and compare against the zero-shot canonical baseline.}
    \label{fig:locked_joints}
\end{figure*}

We first evaluate the overall effectiveness of DexFormer under the standard training setting with 5-step history. All policies are trained with 4096 parallel environments episodes and evaluated on both \textit{zero-shot generalization} on the canonical hand and a 32-variant test set across LEAP, Allegro, and RAPID hands. For each evaluation, we execute the policy across 32 parallel environments for a total of 100 episodes.

As shown in TABLE~\ref{tab:backbone}, DexFormer consistently outperforms LSTM- and GRU-based baselines in grasping success rate after 40k training episodes, addressing Q1. Moreover, it demonstrates strong zero-shot generalization across heterogeneous robotic hands, addressing Q2.

\begin{table}[h!]
\centering
\caption{Comparison with LSTM and GRU-based baselines.}
\label{tab:backbone}
\begin{tabular}{llccc}
\toprule
Hand & Setting  & LSTM & GRU & Ours \\
\midrule
\multirow{2}{*}{{LEAP}}
  & canonical       & 66.81 & 58.91 & \textbf{83.25} \\
  & 32 variants    & 66.72 & 57.38 & \textbf{86.84} \\
\midrule
\multirow{2}{*}{{Allegro}}
  & canonical       & 65.38 & 25.81 & \textbf{74.19} \\
  & 32 variants     & 62.44 & 24.97 & \textbf{71.94} \\
\midrule
\multirow{2}{*}{{RAPID}}
  & canonical       & 46.72 & 45.06 & \textbf{71.69} \\
  & 32 variants     & 44.22 & 53.59 & \textbf{77.09} \\
\midrule
\textbf{Average} & Combined & 58.72 &  44.29  & \textbf{77.50} \\
\bottomrule
\end{tabular}
\end{table}

To answer Q3, we assess the effect of temporal context by comparing DexFormer policies with a single history step and  with a 5-step history. As shown in Table~\ref{tab:main_history_length}, under an equal 40k-episode training budget, incorporating observation–action histories improves zero-shot performance for both the LEAP and Allegro hands. 

\begin{table}[h!]
\centering
\caption{Comparison of history length.}
\label{tab:main_history_length}
\begin{tabular}{llcc}
\toprule
Hand & Setting & 1-step & 5-step (Ours) \\
\midrule
\multirow{2}{*}{{LEAP}}
  & canonical     & 82.72 & \textbf{83.25} \\
  & 32 variants   & 83.34 & \textbf{86.84} \\
\midrule
\multirow{2}{*}{{Allegro}}
  & canonical     & 56.47 & \textbf{74.19} \\
  & 32 variants   & 55.16 & \textbf{71.94} \\
\midrule
\multirow{2}{*}{{RAPID}}
  & canonical     & \textbf{77.59} & 71.69 \\
  & 32 variants   & \textbf{82.75} & 77.09 \\
\midrule
\textbf{Average} & Combined & 73.00 & \textbf{77.50} \\
\bottomrule
\end{tabular}
\end{table}
We study the effect of morphological diversity during training by varying the number of distinct training embodiments used. For this experiment, we trained three policies and used single GPU per each hand type. As summarized in Table~\ref{tab:ablation_urdf_num}, for a 25k-episode training, increasing embodiment diversity substantially improves zero-shot grasping performance on unseen hands, indicating that exposure to heterogeneous embodiments during training enhances cross-embodiment generalization, addressing Q4.

\begin{table}[h!]
\centering
\caption{Ablation on training set embodiment diversity.}
\label{tab:ablation_urdf_num}
\begin{tabular}{llccc}
\toprule
Hand & Setting & 25 Hands & 50 Hands & 100 Hands (Ours) \\
\midrule
\multirow{2}{*}{{LEAP}}
  & canonical     & 81.59 & 80.28 & \textbf{90.12} \\
  & 32 variants   & 81.44 & 84.59 & \textbf{91.31} \\
\midrule
\multirow{2}{*}{{Allegro}}
  & canonical     & 80.34 & 77.84 & \textbf{80.81} \\
  & 32 variants   & 77.12 & 74.44 & \textbf{78.38} \\
\midrule
\multirow{2}{*}{{RAPID}}
  & canonical     &    82.09   & 80.00 & \textbf{84.56} \\
  & 32 variants   &    78.47   & 78.88 & \textbf{79.41} \\
\midrule
\textbf{Average} & Combined & 80.18 & 79.34 & \textbf{84.09}\\
\bottomrule
\end{tabular}
\end{table}
To answer Q5, we further analyze the scalability of DexFormer with respect to environment-level parallelism. As shown in Table~\ref{tab:ablation_env_num}, under a 30k-episode training budget and using 3 GPUs for distributed training, increasing the number of parallel environments per GPU improves performance for LEAP and Allegro hands. However, we noticed some performance drop for zero-shot transfer to the canonical RAPID hand and its variants, which we attribute to an imbalance in morphology statistics during training.

\begin{table}[h!]
\centering
\caption{Ablation on number of environments in parallel.}
\label{tab:ablation_env_num}
\begin{tabular}{llccc}
\toprule
Hand & Setting & 1024/GPU & 2048/GPU & 4096/GPU (Ours) \\
\midrule
\multirow{2}{*}{{LEAP}}
  & canonical     & 66.22 & 82.53 & \textbf{87.00} \\
  & 32 variants   & 69.03 & 80.66 & \textbf{88.78} \\
\midrule
\multirow{2}{*}{{Allegro}}
  & canonical     & 67.16 & 70.78 & \textbf{77.50} \\
  & 32 variants   & 63.94 & 69.12 & \textbf{77.12} \\
\midrule
\multirow{2}{*}{{RAPID}}
  & canonical     & \textbf{74.34} & 69.91 & 69.72 \\
  & 32 variants   & \textbf{79.56} & 76.97 & 77.72 \\
  \midrule
\multirow{2}{*}{\textbf{Average}}
  & canonical     & 69.24 & 74.41 & \textbf{78.07} \\
  & 32 variants   & 70.84 & 75.58 & \textbf{81.21} \\
\bottomrule
\end{tabular}
\end{table}

Locking individual joints induces structured degradation patterns that reflect the anatomical role and kinematic importance of each DoF, shown in Fig.~\ref{fig:locked_joints}. We evaluate using the optimal DexFormer checkpoint reported in Table~\ref{tab:backbone}. Across all embodiments, distal flexion joints (PIP/DIP) and low-leverage MCP abduction joints cause mild reductions, while basal flexion and thumb CMC joints produce pronounced drops, consistent with their contribution to establishing stable contact and opposition. Notably, all impaired variants achieve success rates that are lower than, or at most match, the canonical hand’s zero-shot performance, indicating no compensatory gain from joint removal. The consistent gap between 2-s and 4-s execution windows suggests that extended rollout enables partial compensation even under impaired morphology, addressing Q6.

\subsection{Real-world evaluation}

\begin{figure}[ht]
    \centering
    \includegraphics[width=0.325\linewidth]{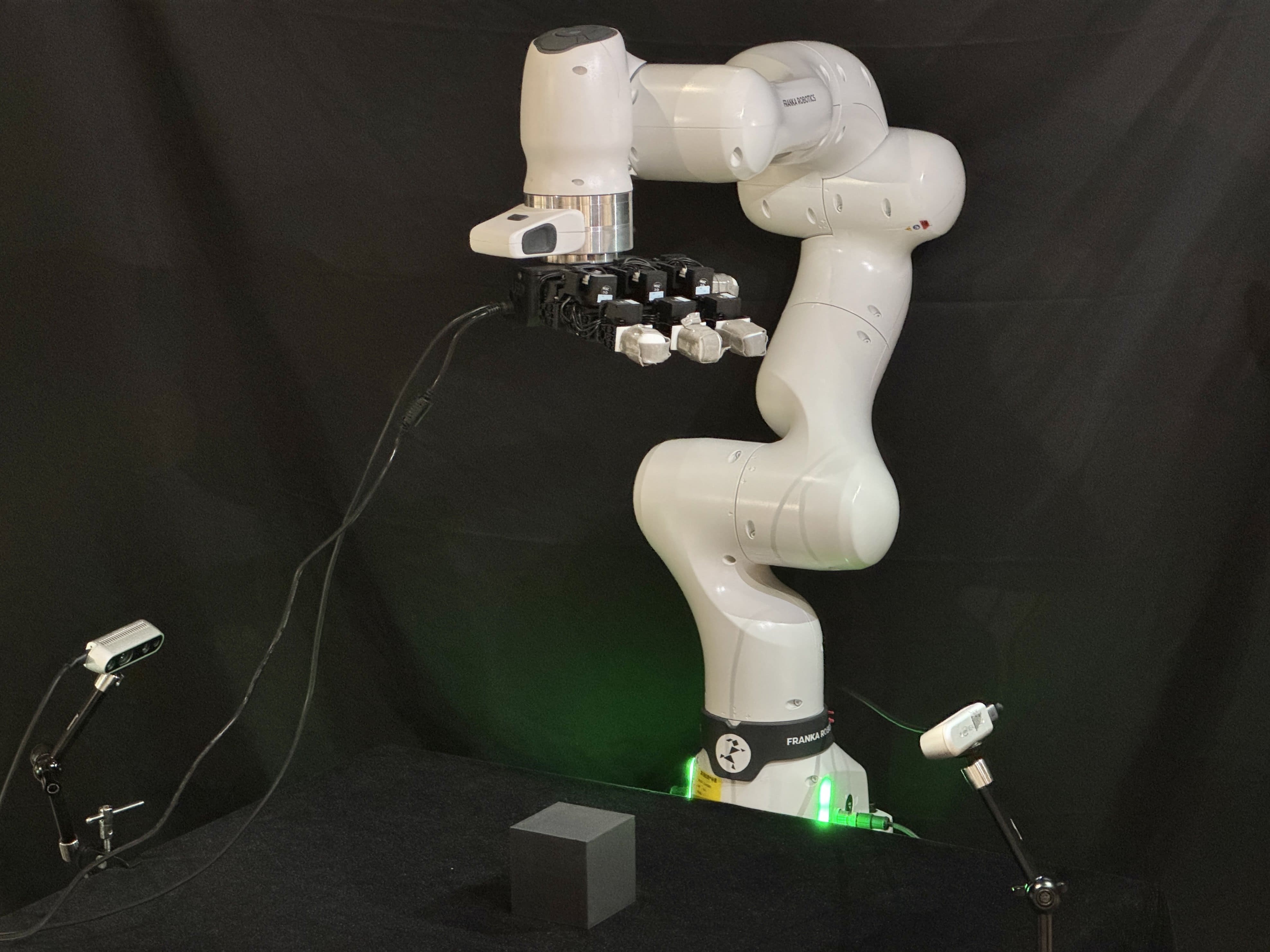}
    \includegraphics[width=0.325\linewidth]{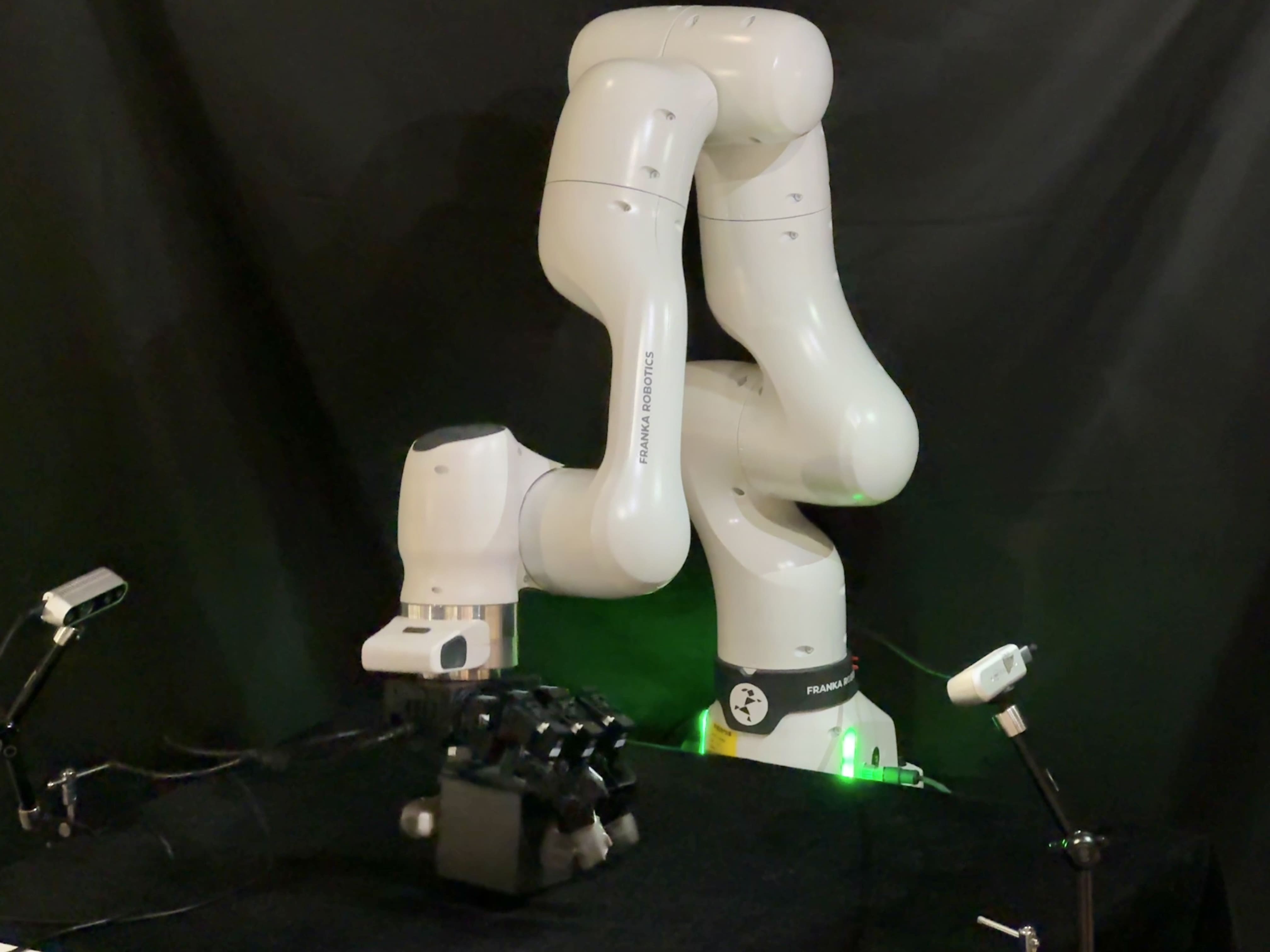}
    \includegraphics[width=0.325\linewidth]{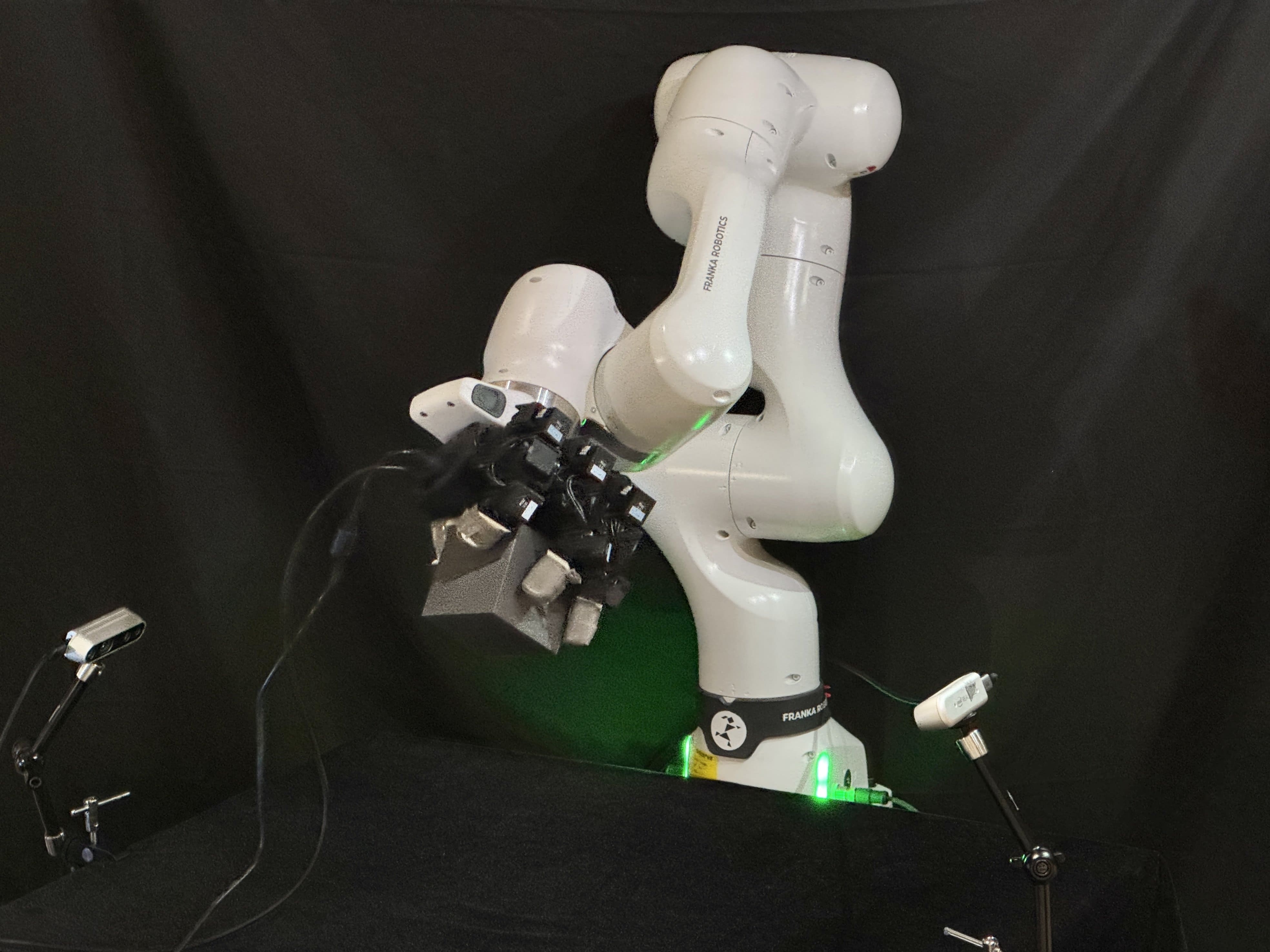}
    \caption{Real-world evaluation of the DexFormer policy.}
    \label{fig:realworld}
\end{figure}

We use a LEAP hand mounted on Franka arm for real-world experiments. To maintain consistency with the simulated ray-casting observations, the real-world system employs two identically mounted Intel RealSense D435 cameras. The resulting depth measurements are fused through ICP and clipped to the hand workspace to produce a coherent point-cloud representation of both the dexterous hand and manipulated objects. The Franka arm runs its joint-space controller at 1000 Hz, and the learned policy is evaluated at 10 Hz. The real-world setup is shown as in Fig.~\ref{fig:realworld}.

For smooth sim-to-real transfer, we perform offline distillation rather than online distillation, enabling efficient data acquisition and stable training. We recorded 2000 episodes of expert demonstrations with a 158-dimensional observation space. The student MLP network is optimized with Adam using a learning rate of 0.001. 
The deployed policy can successfully lift the object cube into target position.

\section{Conclusion} 
\label{sec:conclusion}
We present DexFormer, a history-conditioned transformer policy for cross-embodiment dexterous manipulation. By conditioning on observation–action histories, DexFormer implicitly infers morphology and dynamics, enabling a single policy to generalize across heterogeneous robotic hands without requiring embodiment identifiers or dedicated decoder heads. Extensive experiments demonstrate that DexFormer (i) outperforms GRU- and LSTM-based baselines, (ii) achieves strong zero-shot transfer across LEAP, Allegro, and RAPID hands and their variants, and (iii) benefits from temporal context, morphology diversity, and environment-level parallelism. These findings provide evidence that dynamics-aware temporal inference is an effective mechanism for cross-embodiment generalization. Real-world evaluations further support the practicality of the approach, indicating that morphology-agnostic dexterous manipulation can extend beyond simulation. Overall, DexFormer offers a scalable path toward unified manipulation policies applicable across diverse embodiments and deployment scenarios.

\section{Limitations}
\label{sec:limitations}

While DexFormer demonstrates strong cross-embodiment generalization for embodiment variations, several limitations remain. First, our current training regime focuses on a limited set of object geometries and mass properties, leading to insufficient object-level generalization. Extending training to larger, more diverse object collections and incorporating multi-expert distillation or mixture-of-experts could improve robustness and transferability to real-world manipulation scenarios. Second, we observe a modest performance drop on higher-DoF embodiments such as RAPID Hand under zero-shot settings. This suggests that more sophisticated architectural or scaling strategies may be beneficial, including deeper temporal modeling, improved action embeddings, and higher degrees of environment-level parallelism. Furthermore, due to current limitations in Isaac Lab, embodiments with distinct kinematic topologies are trained on separate GPUs. Enabling mixed buffer or centralized gradient accumulation across embodiments may reduce gradient variance and improve generalization.

We consider addressing these limitations an important direction for future work, particularly toward foundation-style models for dexterous manipulation. More broadly, we aim to promote a cross-embodiment learning paradigm in which scalable RL training enables high-DoF, morphologically diverse embodiments to be trained at scale, while downstream users can either post-train, finetune, or incorporate residual policies to further adapt to new embodiments and tasks. We view such a path as a step toward more generalized cross-embodiment manipulation policies.

\bibliographystyle{plainnat}
\bibliography{references}

\appendix\label{appendix}

\subsection{Cross-embodiment Training}
\begin{figure}[h!]
    \centering
    \includegraphics[width=0.98\linewidth]{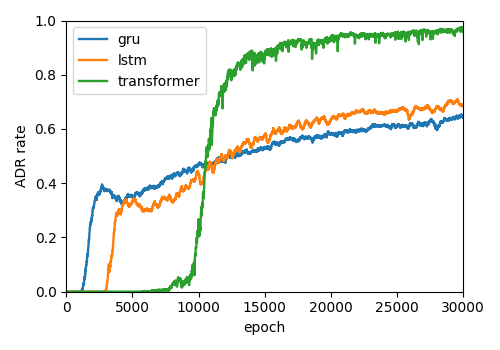}
    \caption{ADR rate with respect to training episodes}
    \label{fig:adr_curve}
\end{figure}

\begin{figure}[h!]
    \centering
    \includegraphics[width=0.98\linewidth]{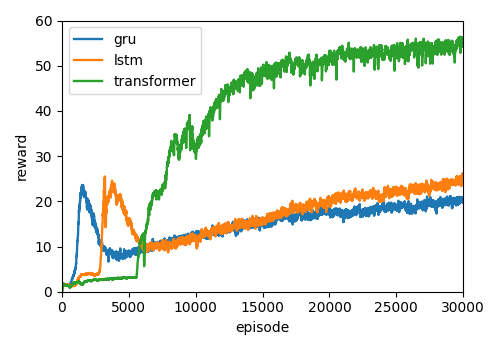}
    \caption{Reward with respect to training episodes}
    \label{fig:reward_curve}
\end{figure}

\begin{figure}[h!]
    \centering
    \includegraphics[width=0.98\linewidth]{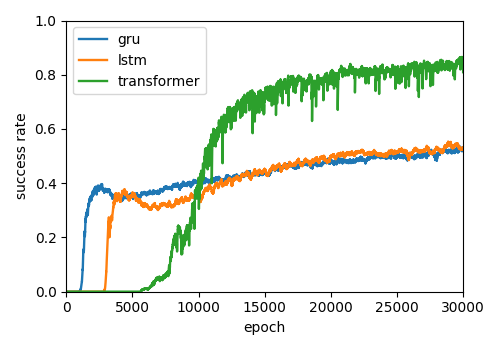}
    \caption{Success rate with respect to training episodes}
    \label{fig:success_curve}
\end{figure}

We employ Automatic Domain Randomization (ADR) to progressively increasing noise levels on joint positions, joint velocities, fingertip states, point-cloud observations, and gravity as the policy performance improves. The ADR curve during training is shown as in Fig.~\ref{fig:adr_curve}.
The learning curves for average reward and success with respect to training episodes are reported in Fig.~\ref{fig:reward_curve} and~\ref{fig:success_curve}. 
As can be seen, our DexFormer policy outperforms LSTM-based and GRU-based policy by training efficiency (curves rise faster) and accuracy.

\subsection{Real-world evaluation}

\begin{figure}[h!]
    \centering
    \includegraphics[width=0.8\linewidth]{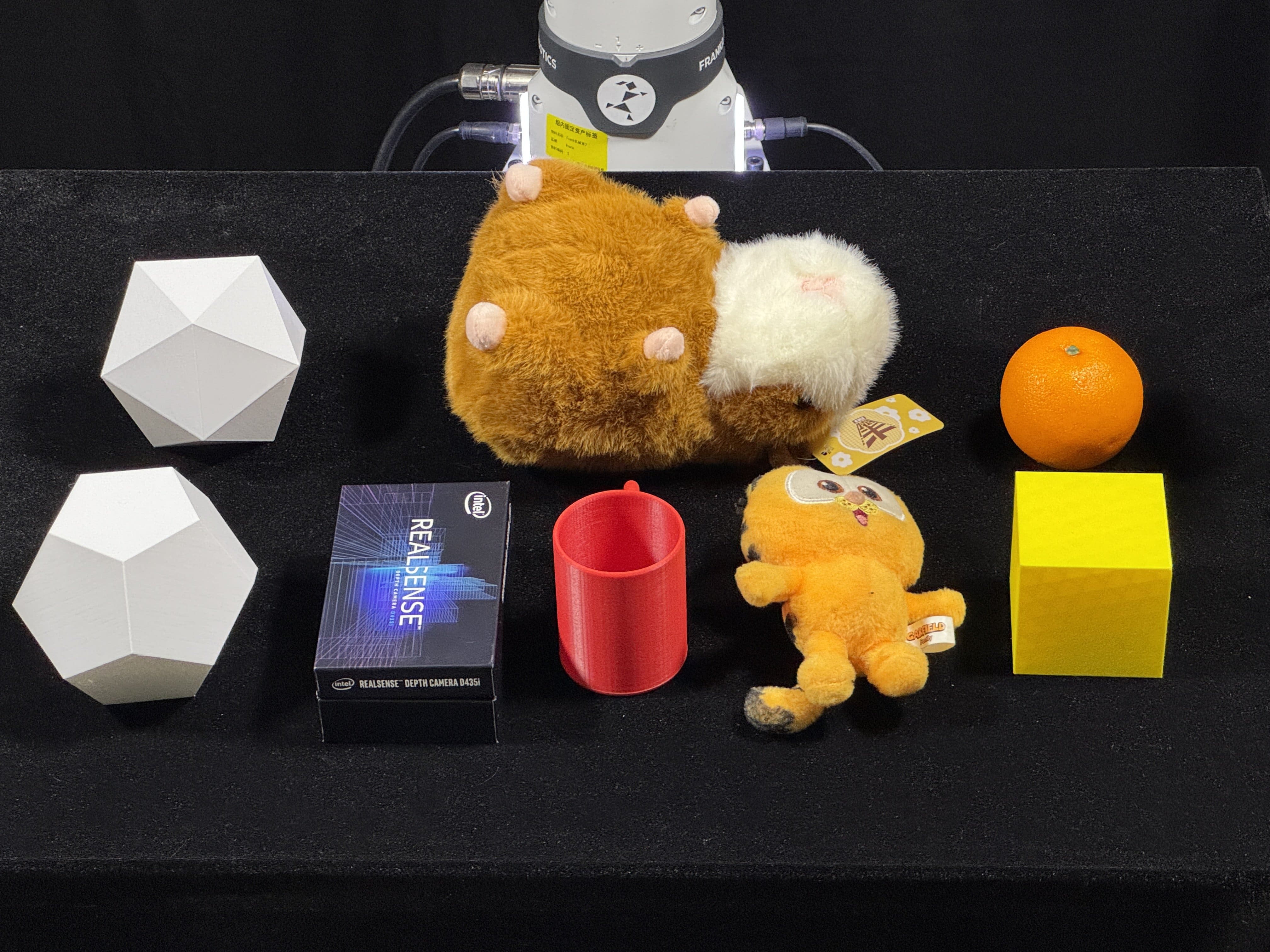}
    \caption{Objects used for real-world experiments}
    \label{fig:real_objects}
\end{figure}

For real world experiments, we use several objects during evaluation: an orange cube, a red mug, a yellow cube, a hamster toy, a cat toy, a dodecahedron, a icosahedron, and a package box, as shown in Fig.~\ref{fig:real_objects}.

\begin{figure*}[h!]
    \centering
    \subfigure[Grasping an orange fruit.]{
        \includegraphics[width=0.15\linewidth]{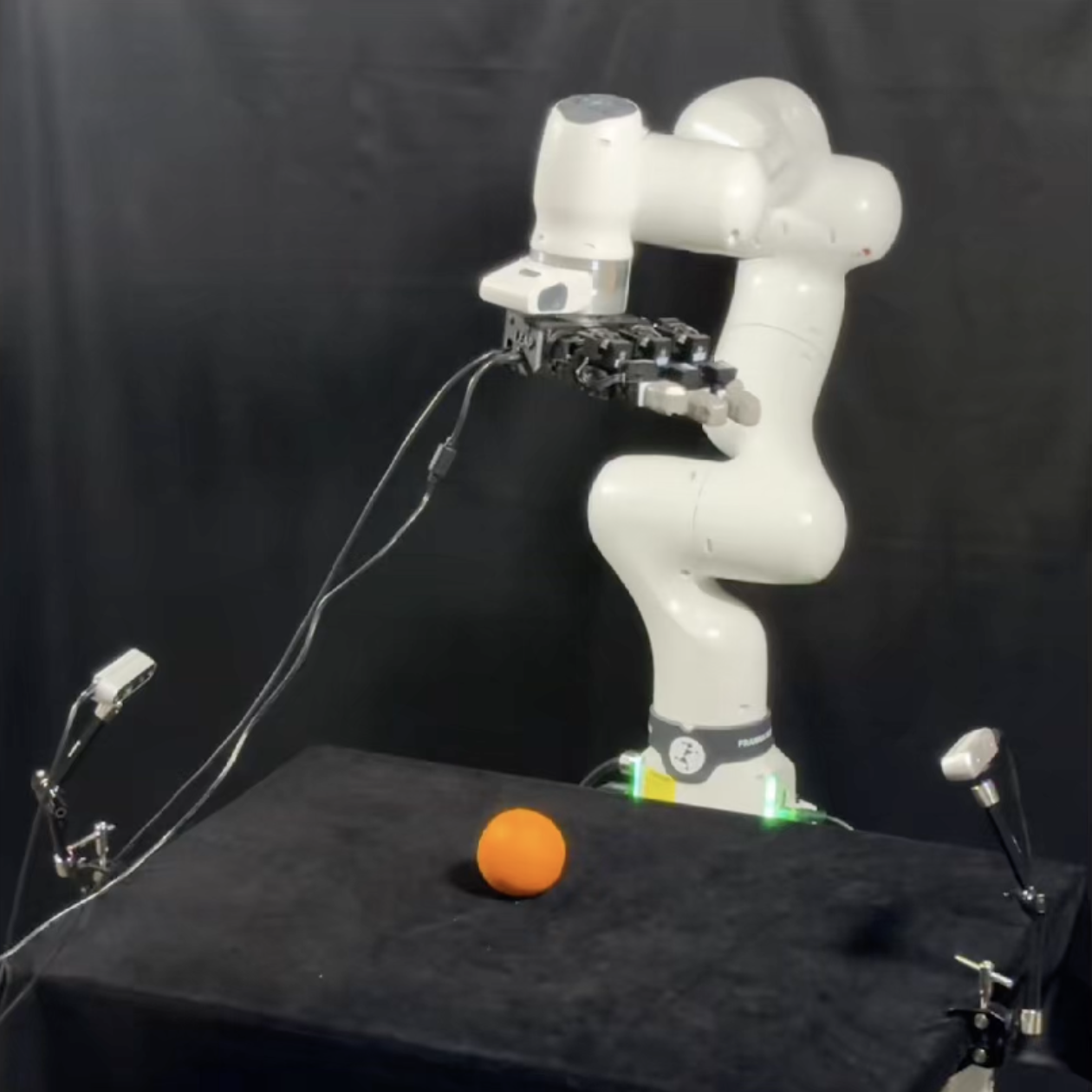}
    \includegraphics[width=0.15\linewidth]{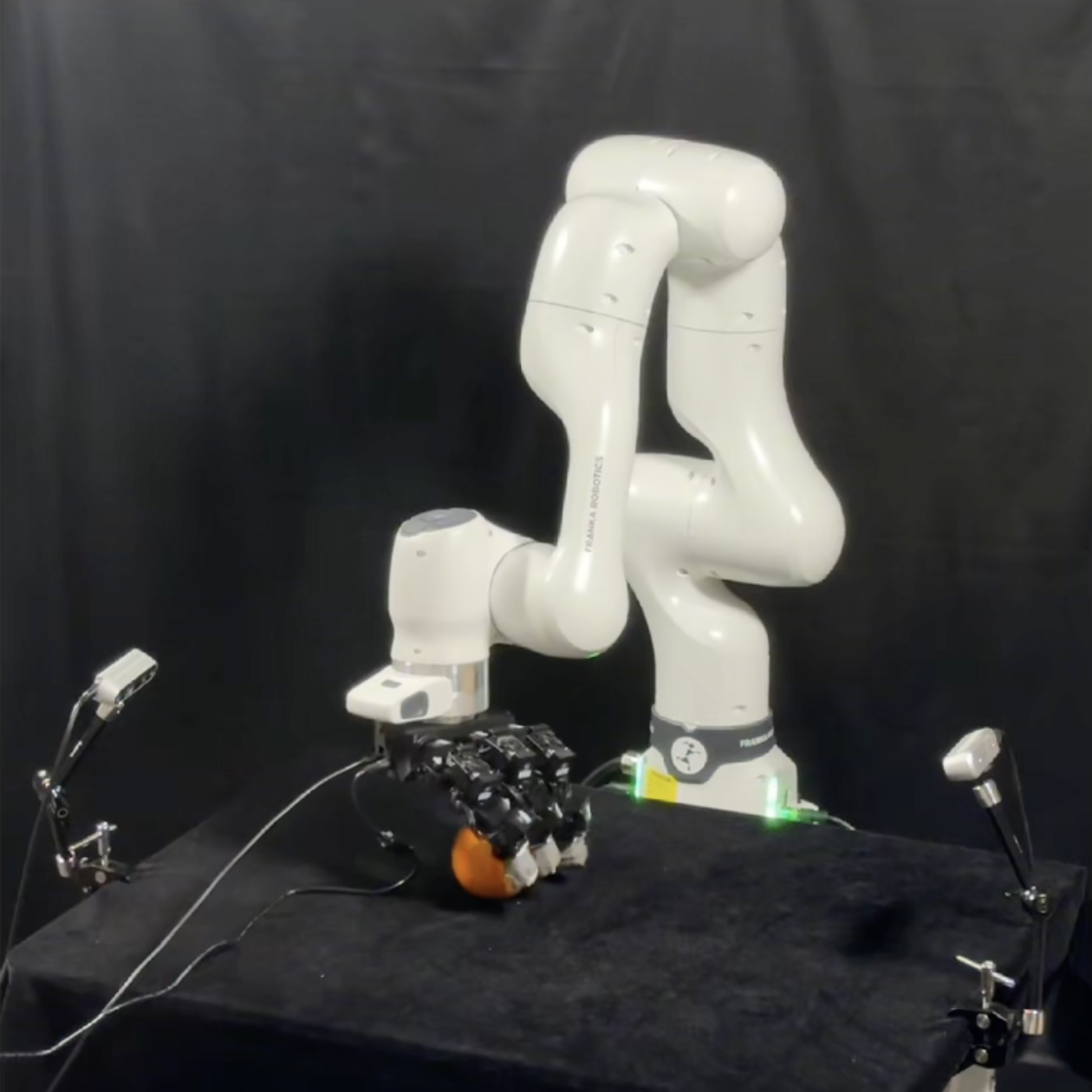}
    \includegraphics[width=0.15\linewidth]{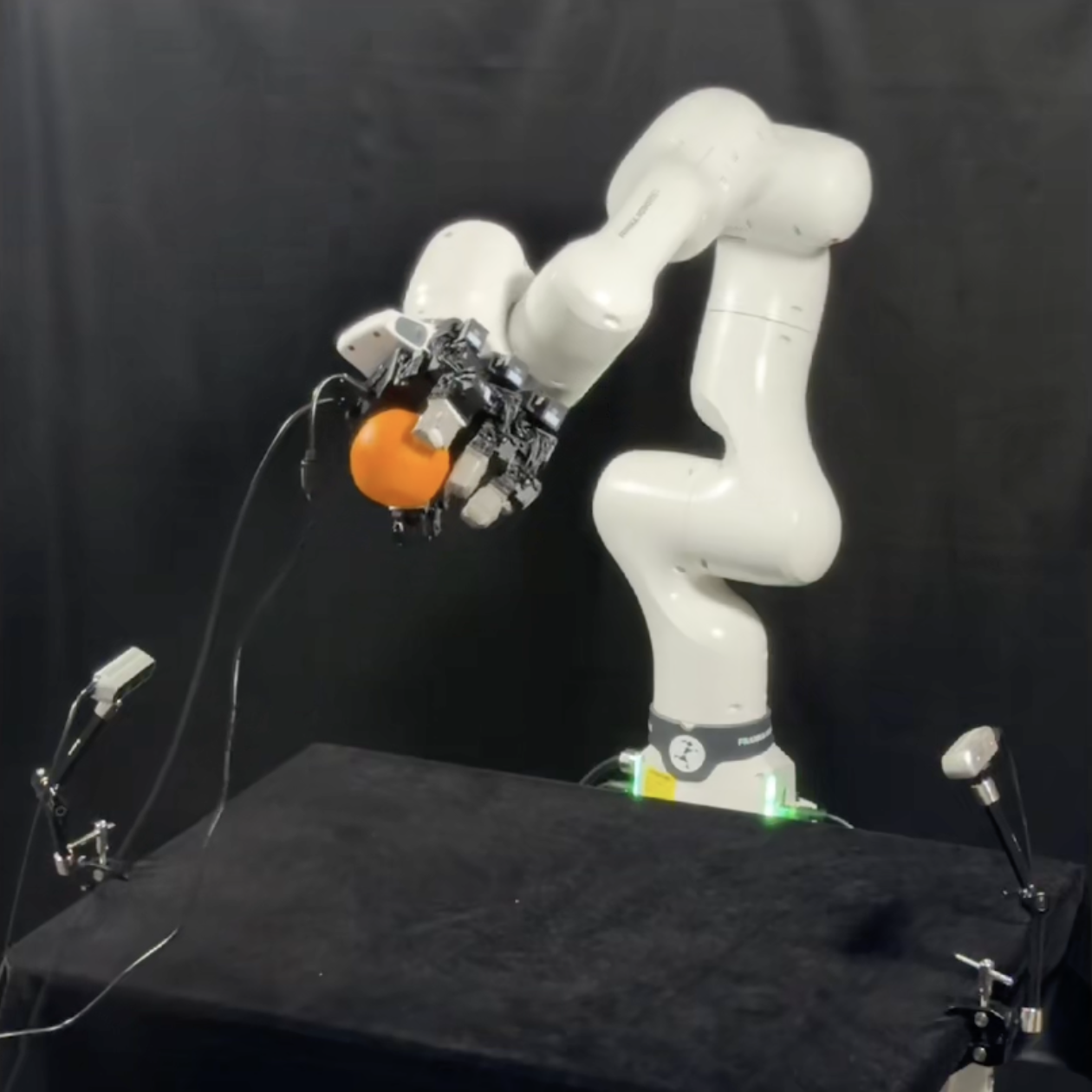}
    \label{fig:orange}
    }
    \subfigure[Grasping a red mug]{
        \includegraphics[width=0.15\linewidth]{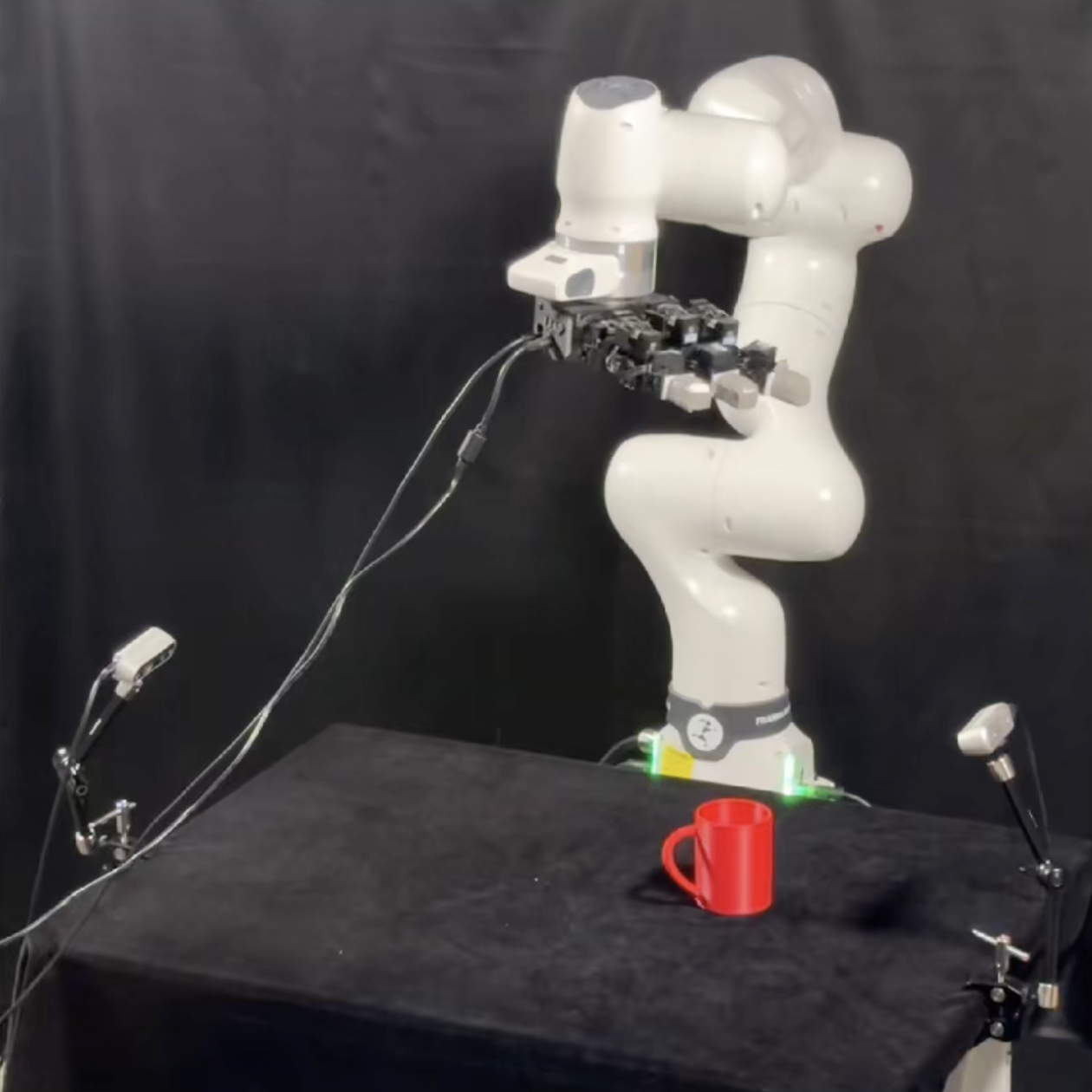}
    \includegraphics[width=0.15\linewidth]{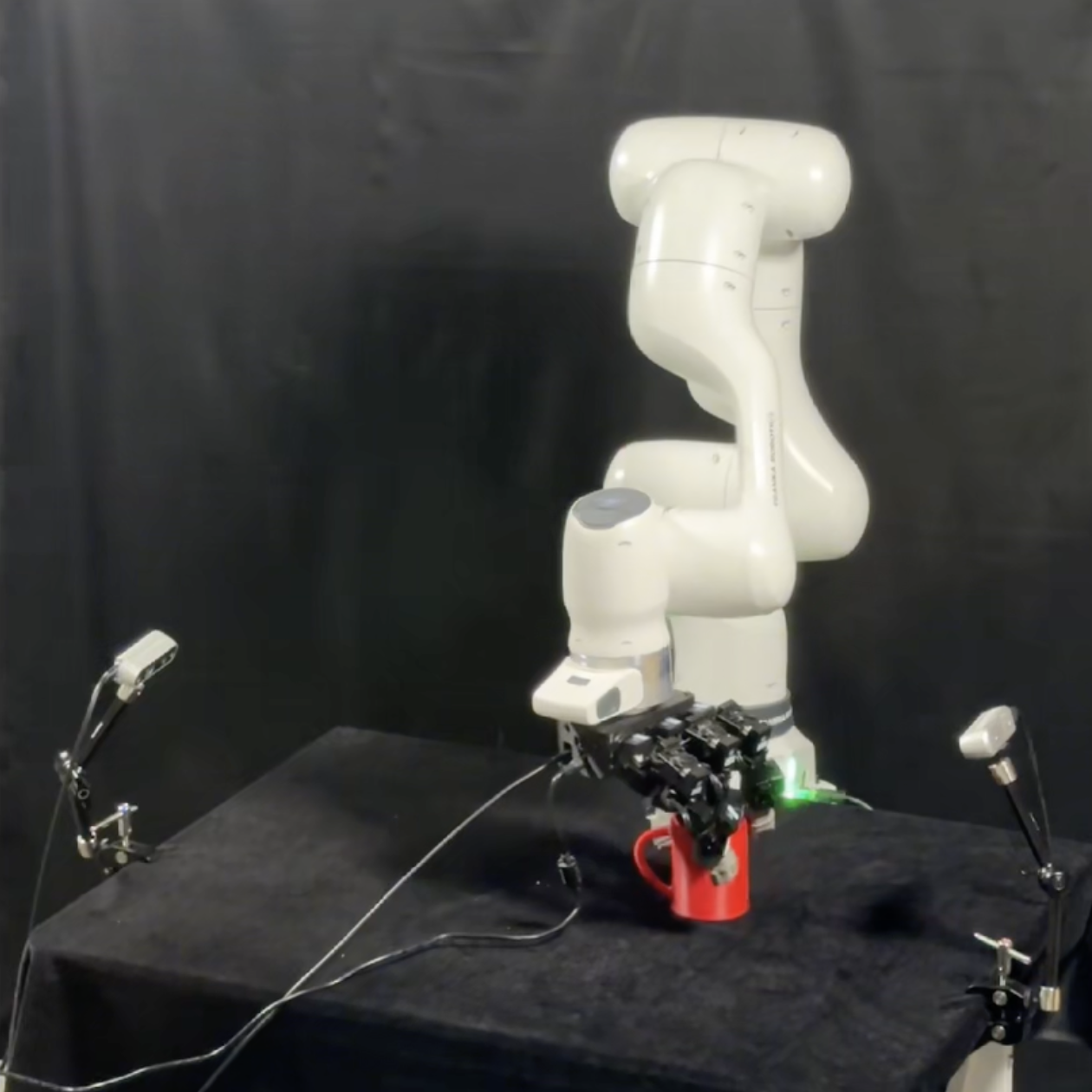}
    \includegraphics[width=0.15\linewidth]{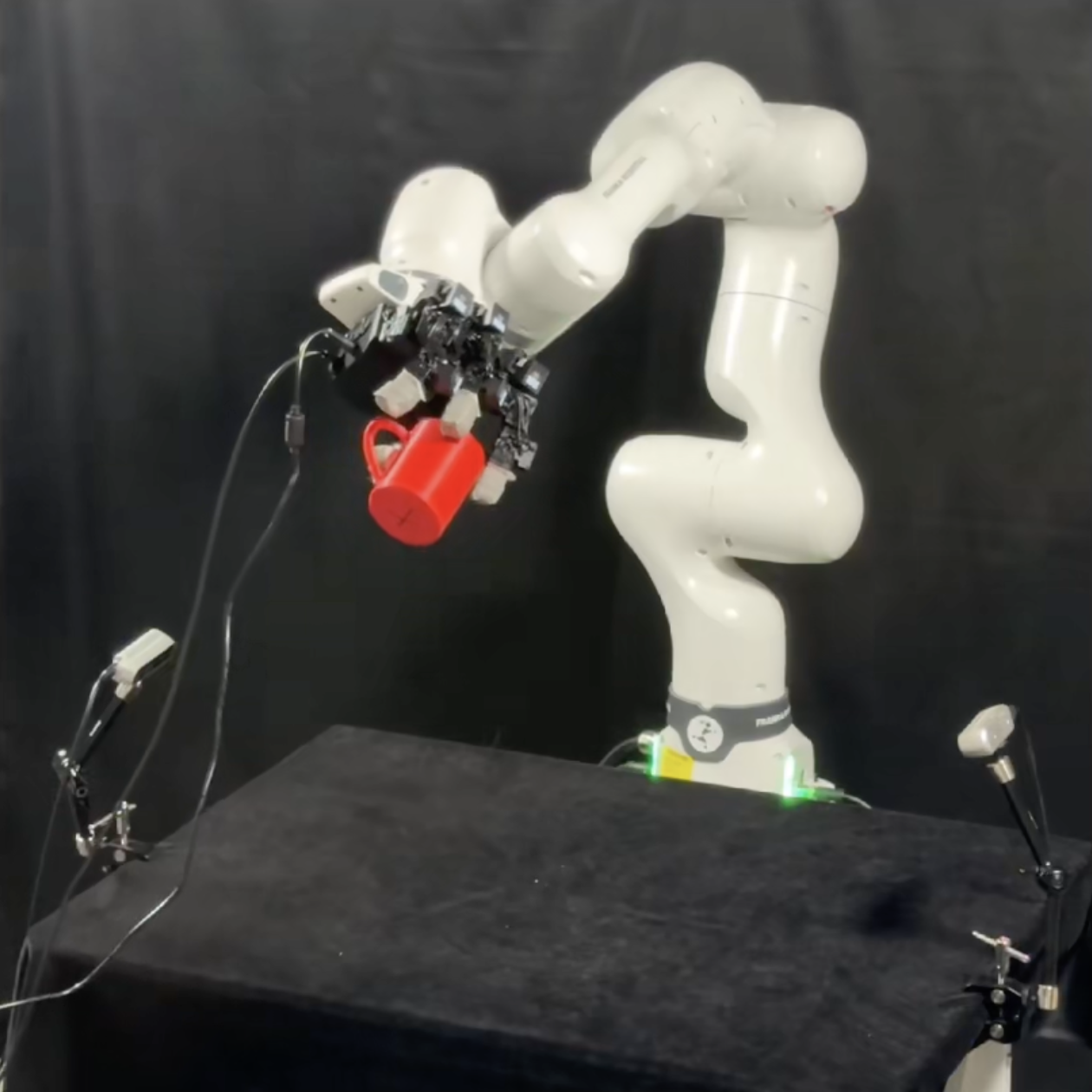}
    \label{fig:mub}
    }
    
    \subfigure[Grasping a yellow cube]{
        \includegraphics[width=0.15\linewidth]{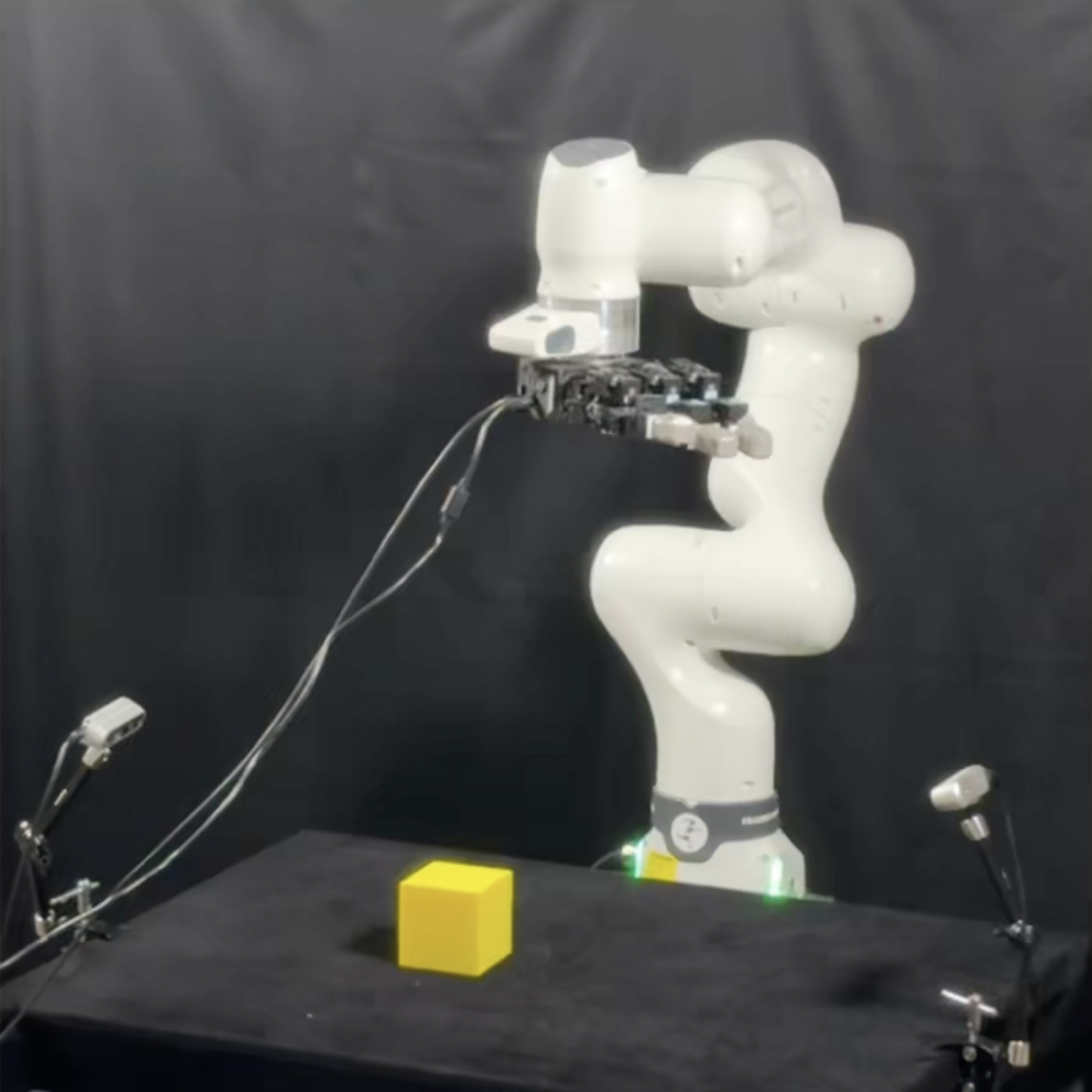}
    \includegraphics[width=0.15\linewidth]{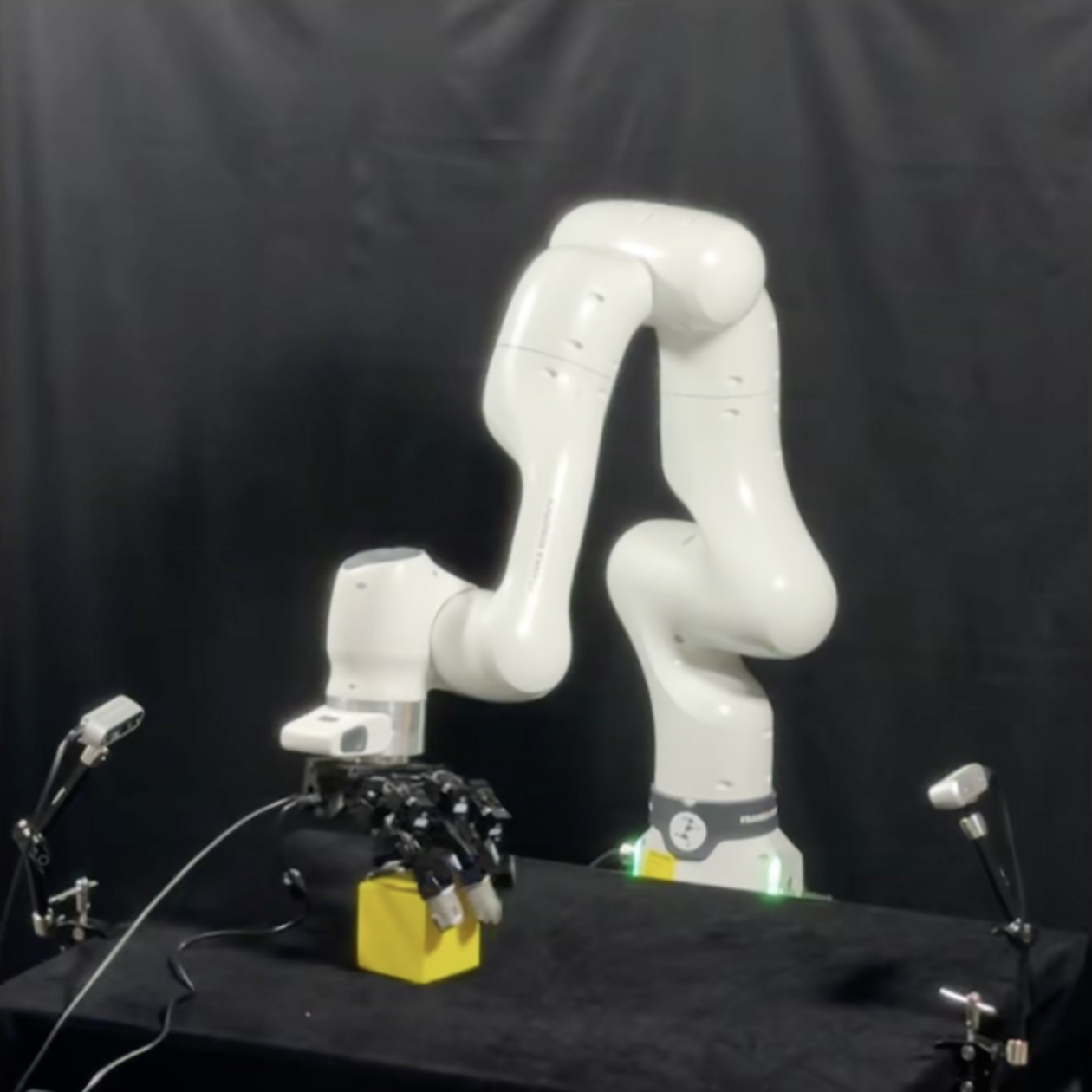}
    \includegraphics[width=0.15\linewidth]{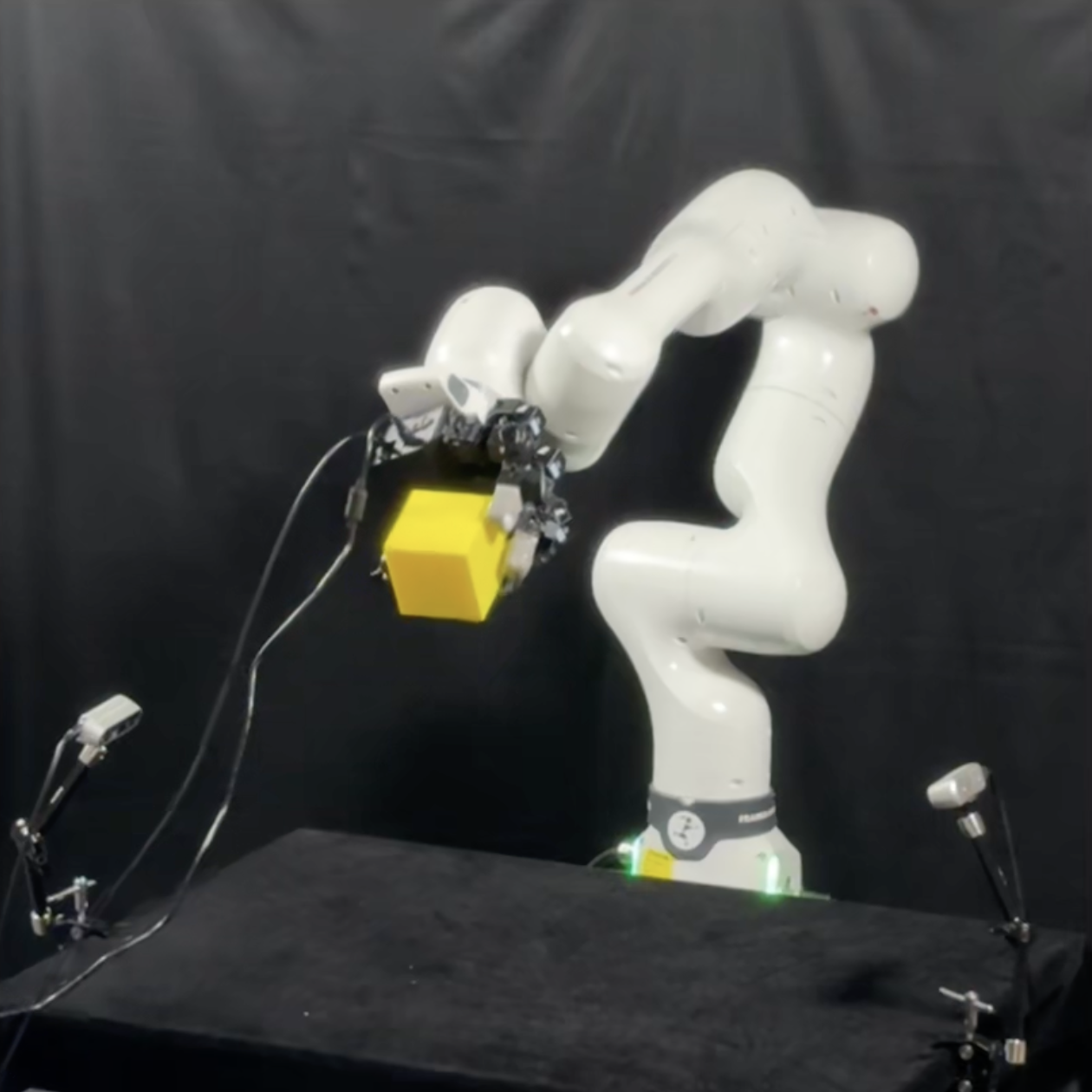}
    \label{fig:cube}
    }
    \subfigure[Grasping a large hamster toy]{
        \includegraphics[width=0.15\linewidth]{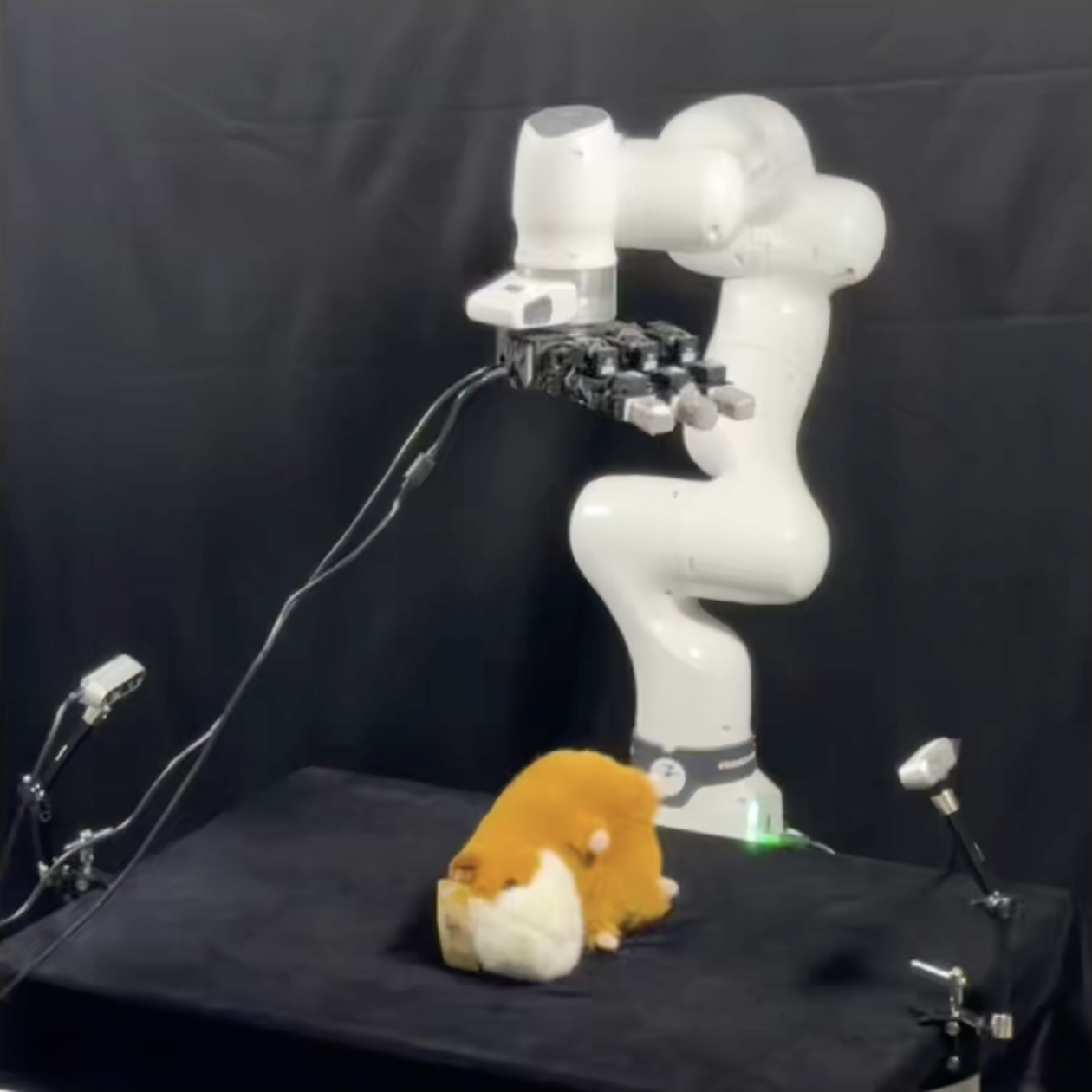}
    \includegraphics[width=0.15\linewidth]{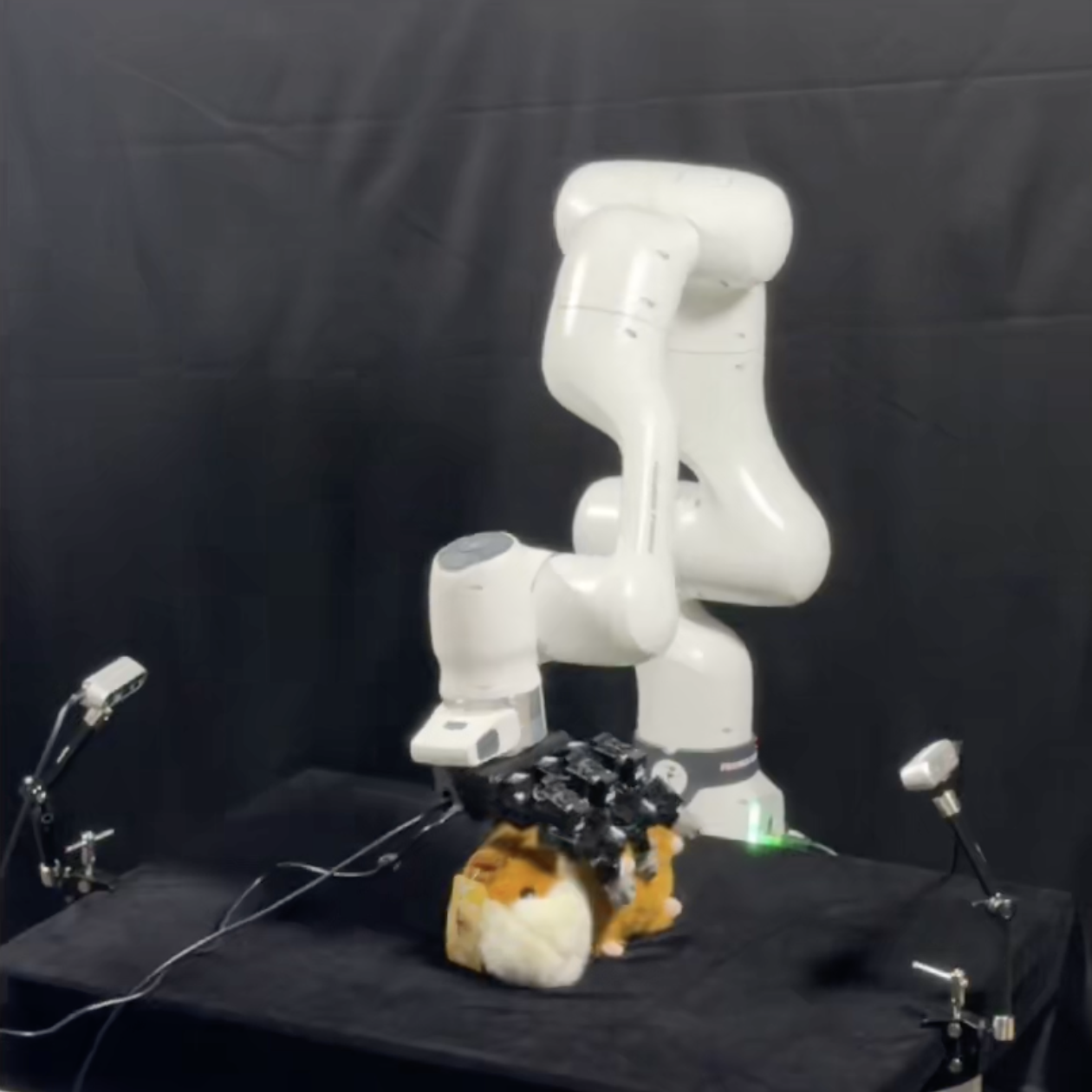}
    \includegraphics[width=0.15\linewidth]{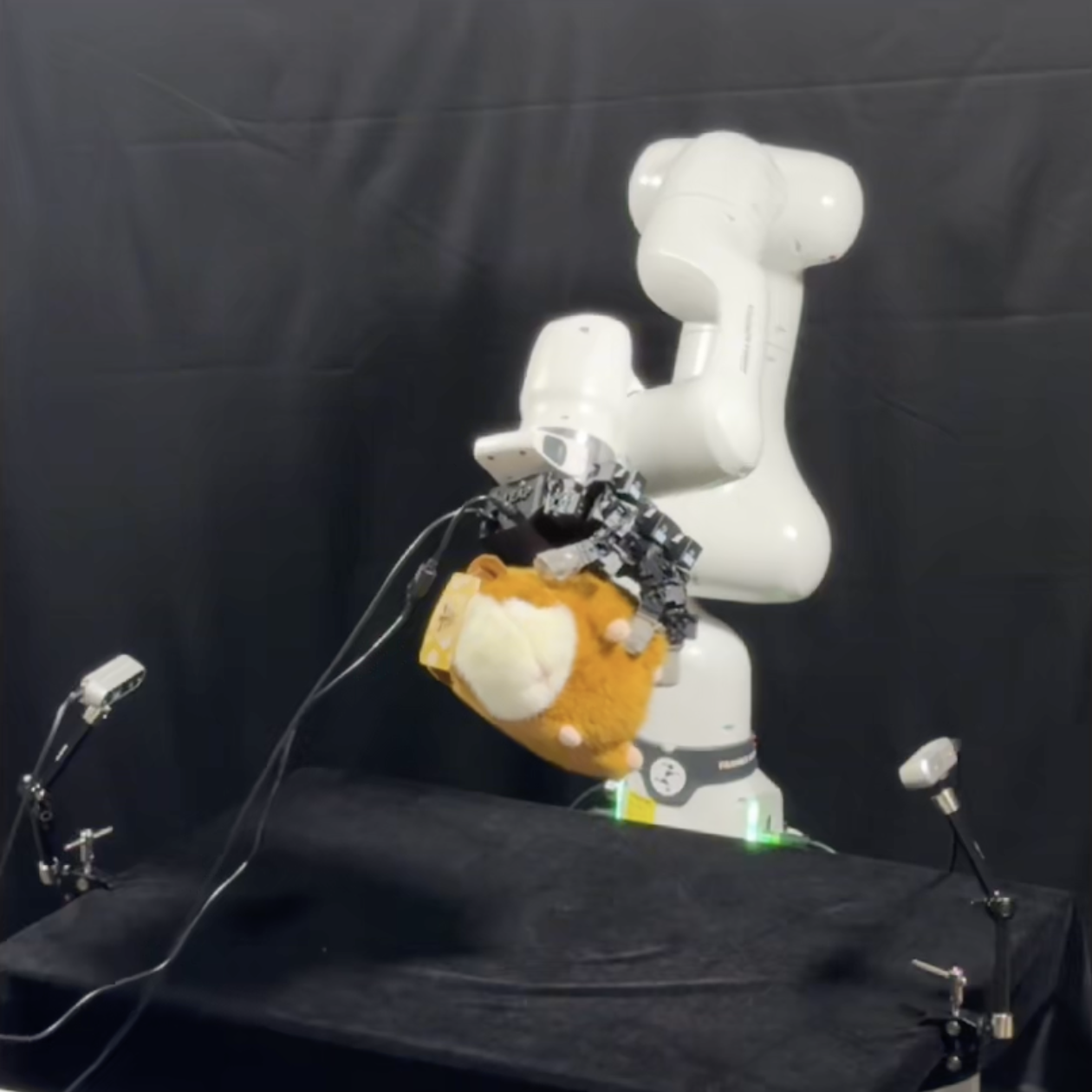}
    \label{fig:hamster}
    }
    
    \subfigure[Grasping a dodecahedron (12-faced polyhedron)]{
        \includegraphics[width=0.15\linewidth]{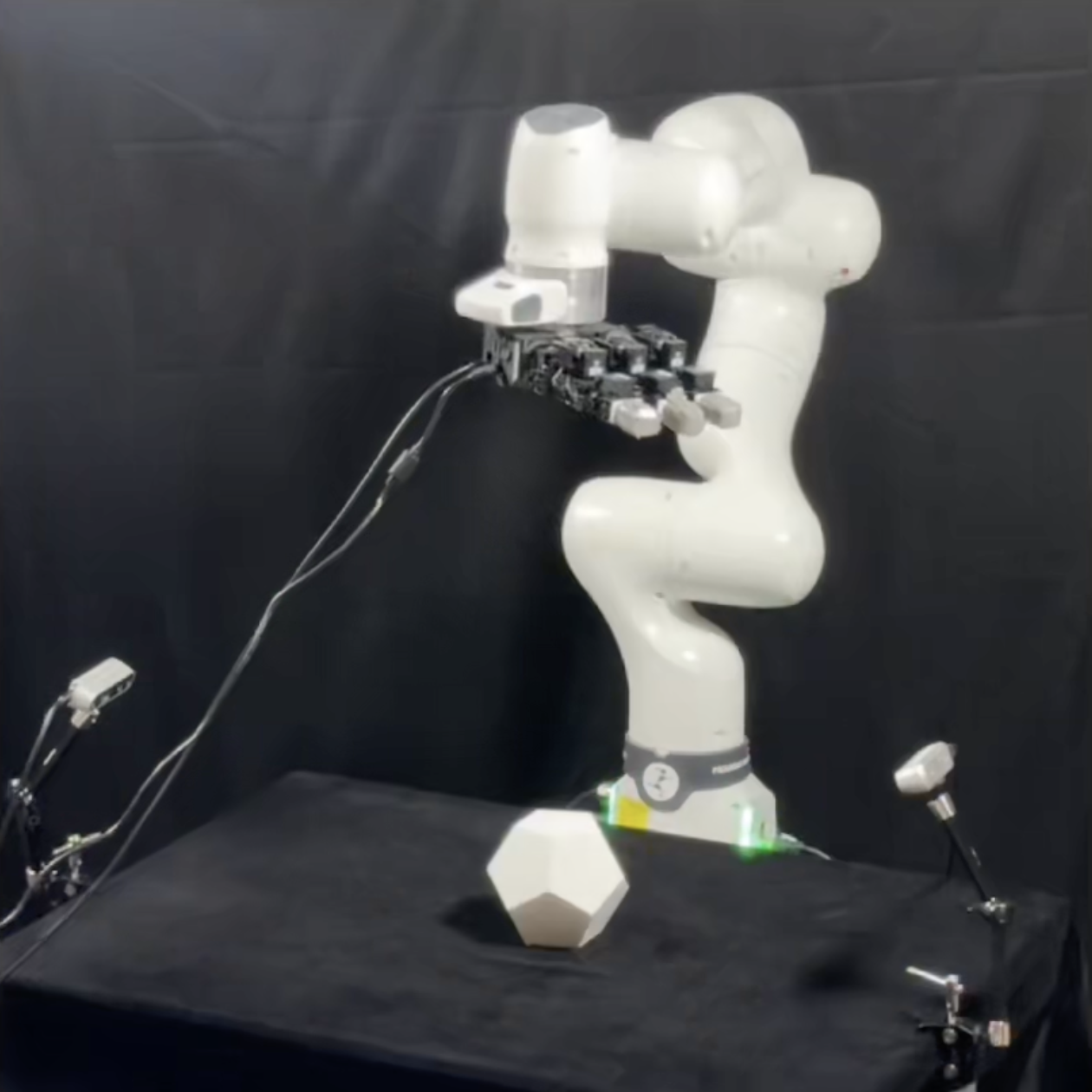}
    \includegraphics[width=0.15\linewidth]{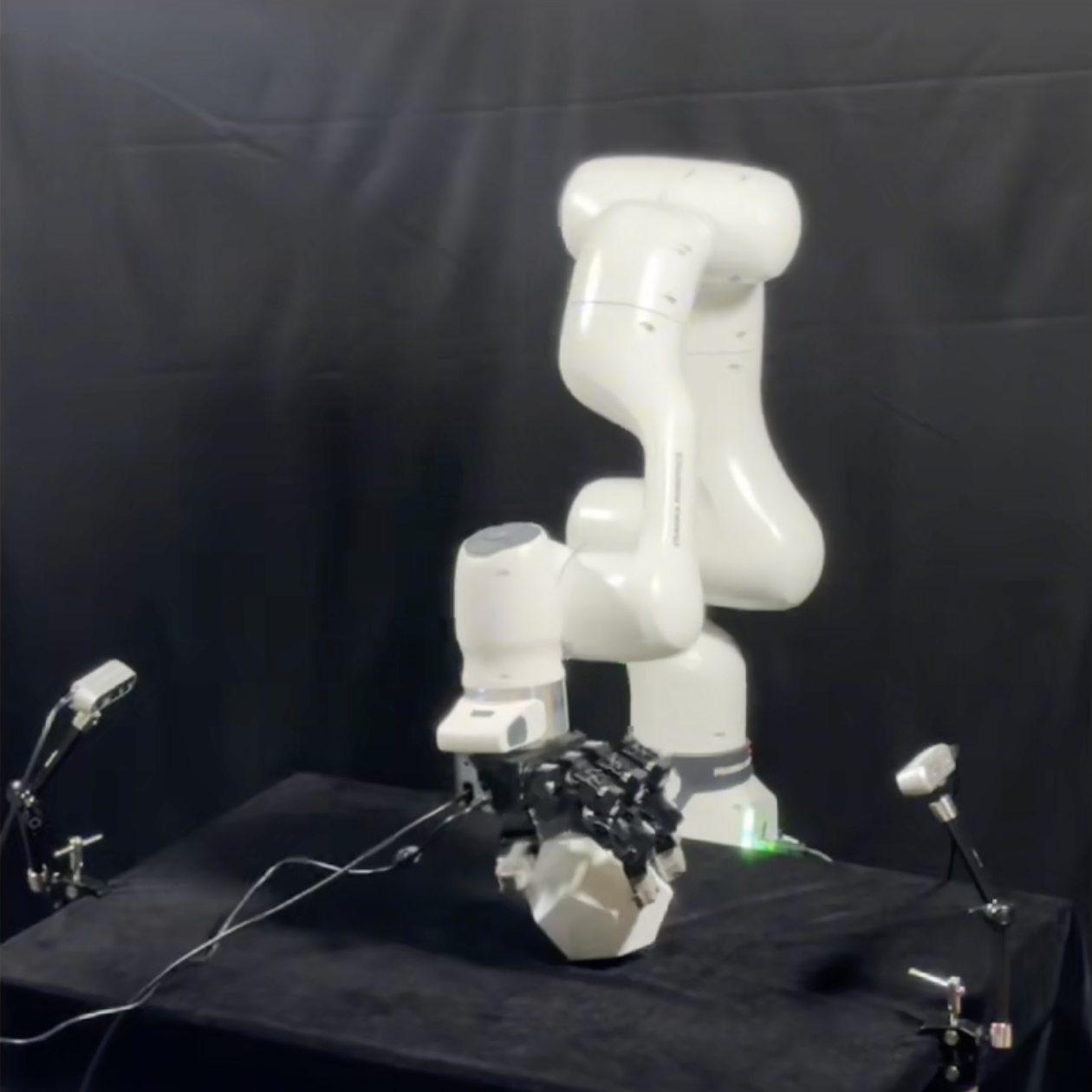}
    \includegraphics[width=0.15\linewidth]{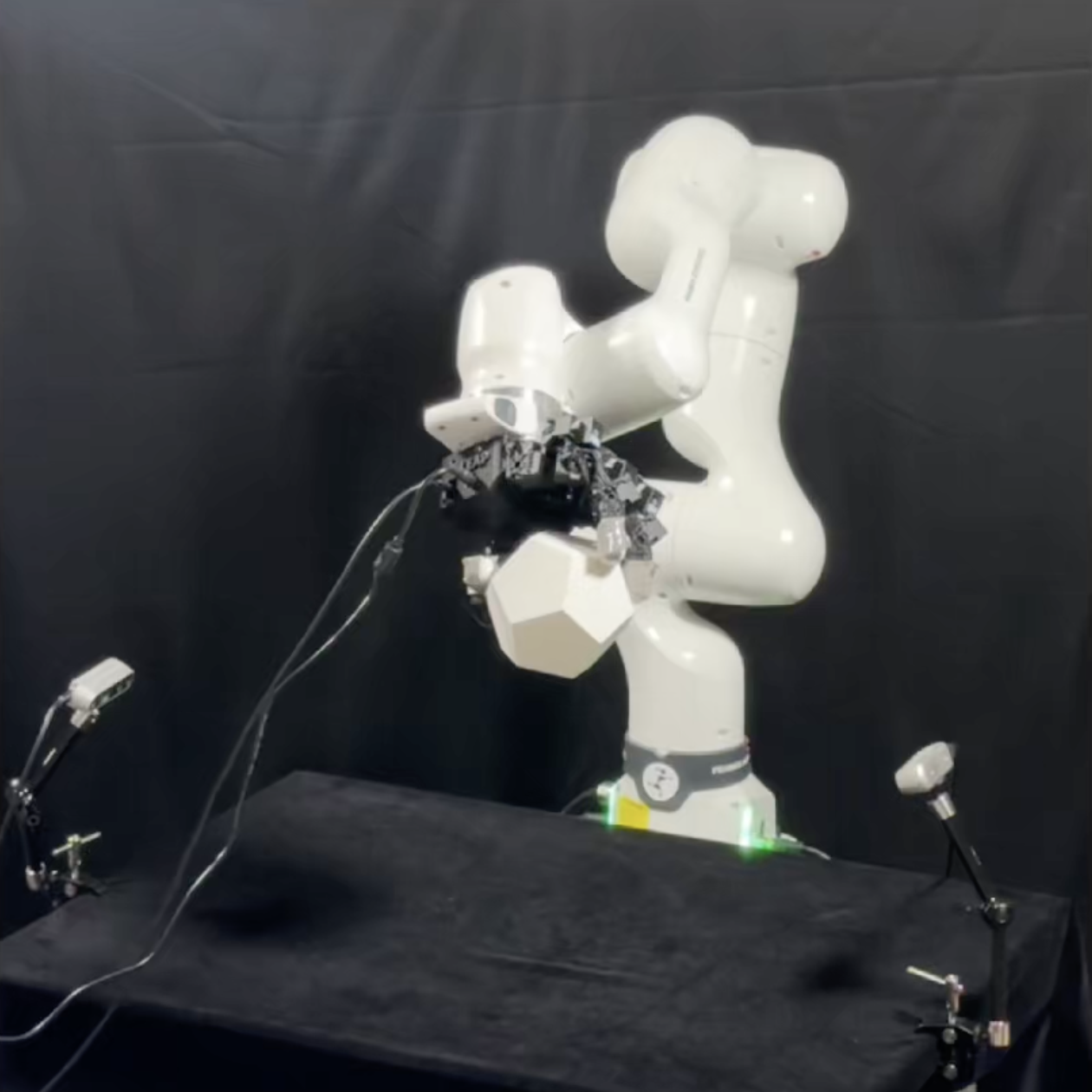}
    \label{fig:dodecahedron}
    }
    \subfigure[Grasping an icosahedron (20-faced polyhedron)]{
        \includegraphics[width=0.15\linewidth]{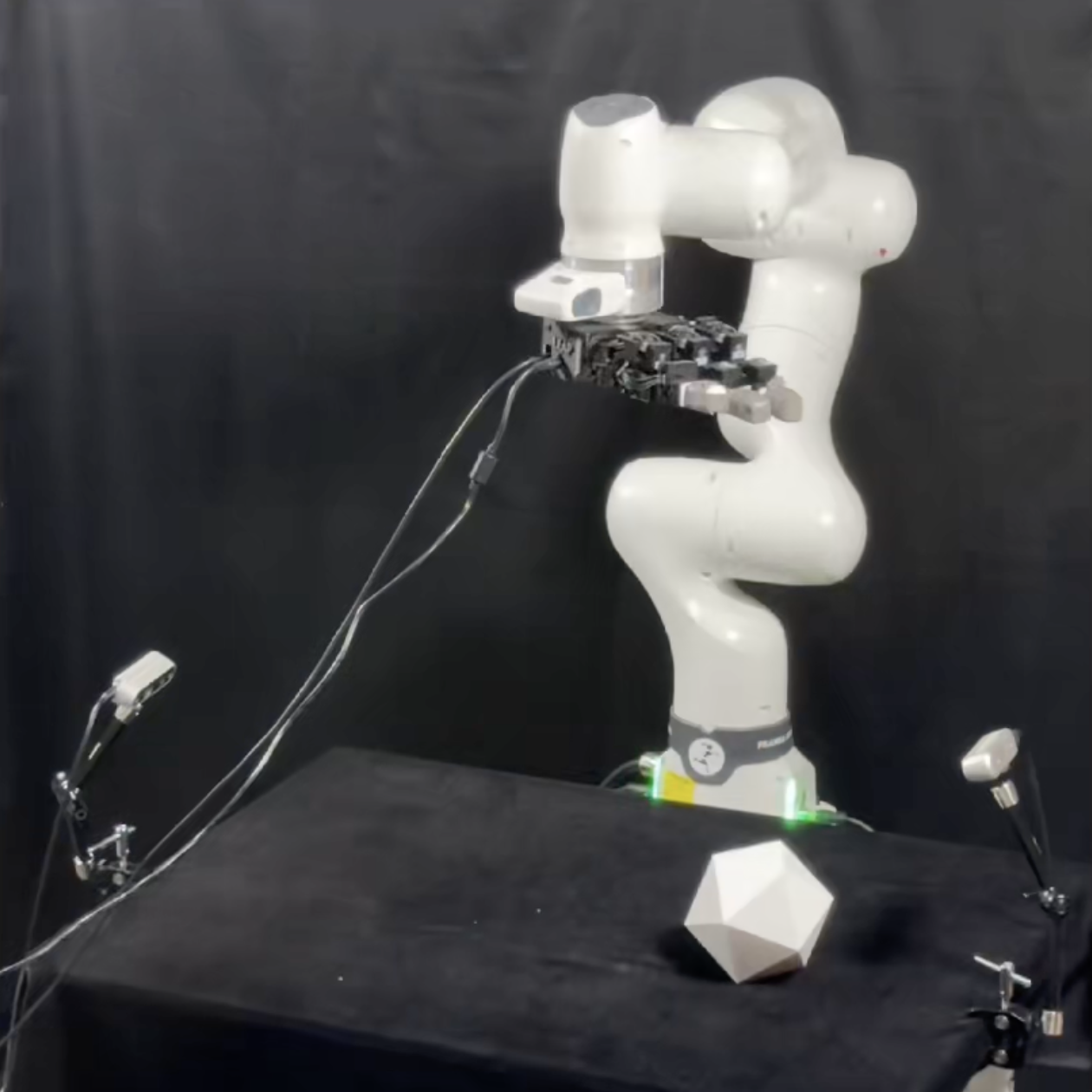}
    \includegraphics[width=0.15\linewidth]{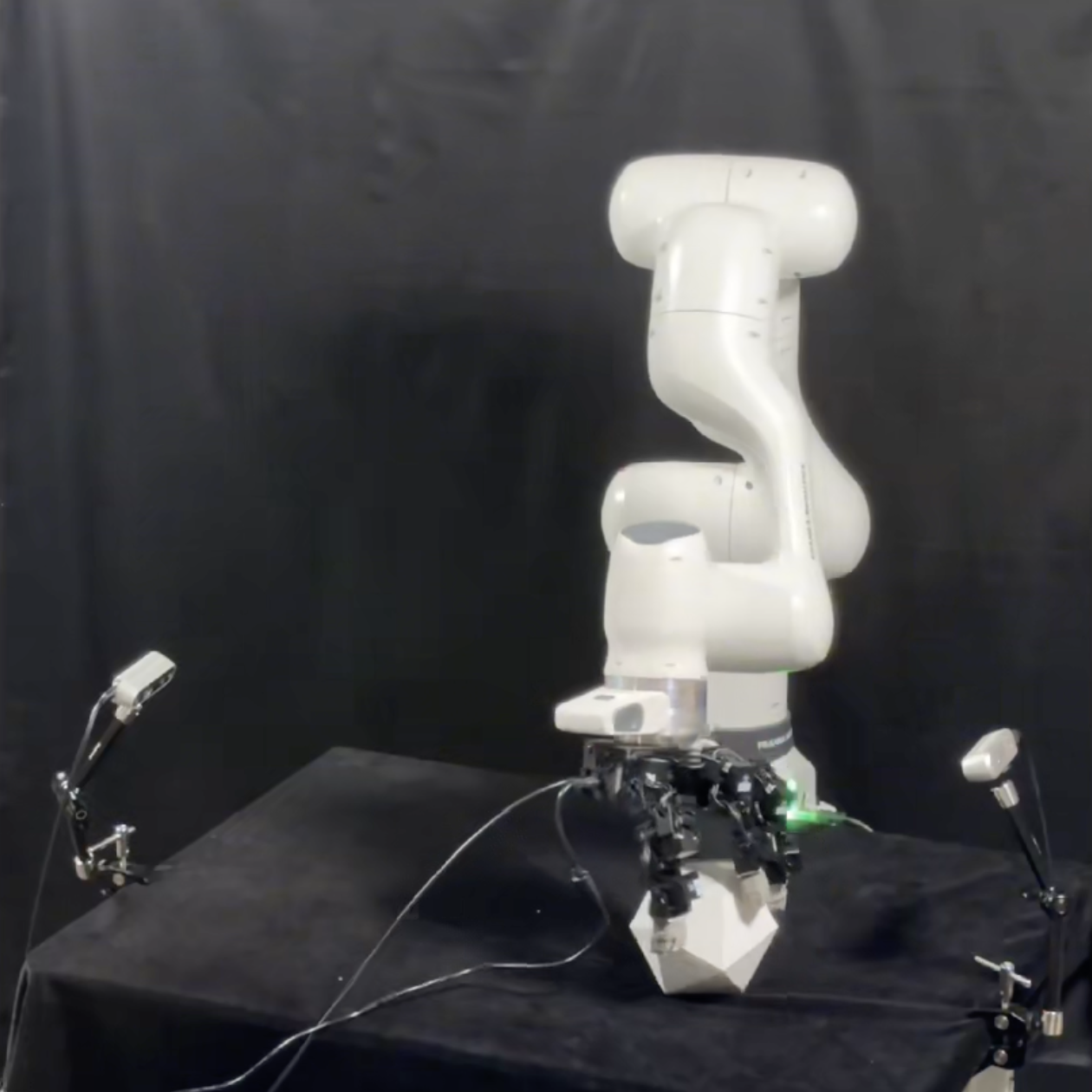}
    \includegraphics[width=0.15\linewidth]{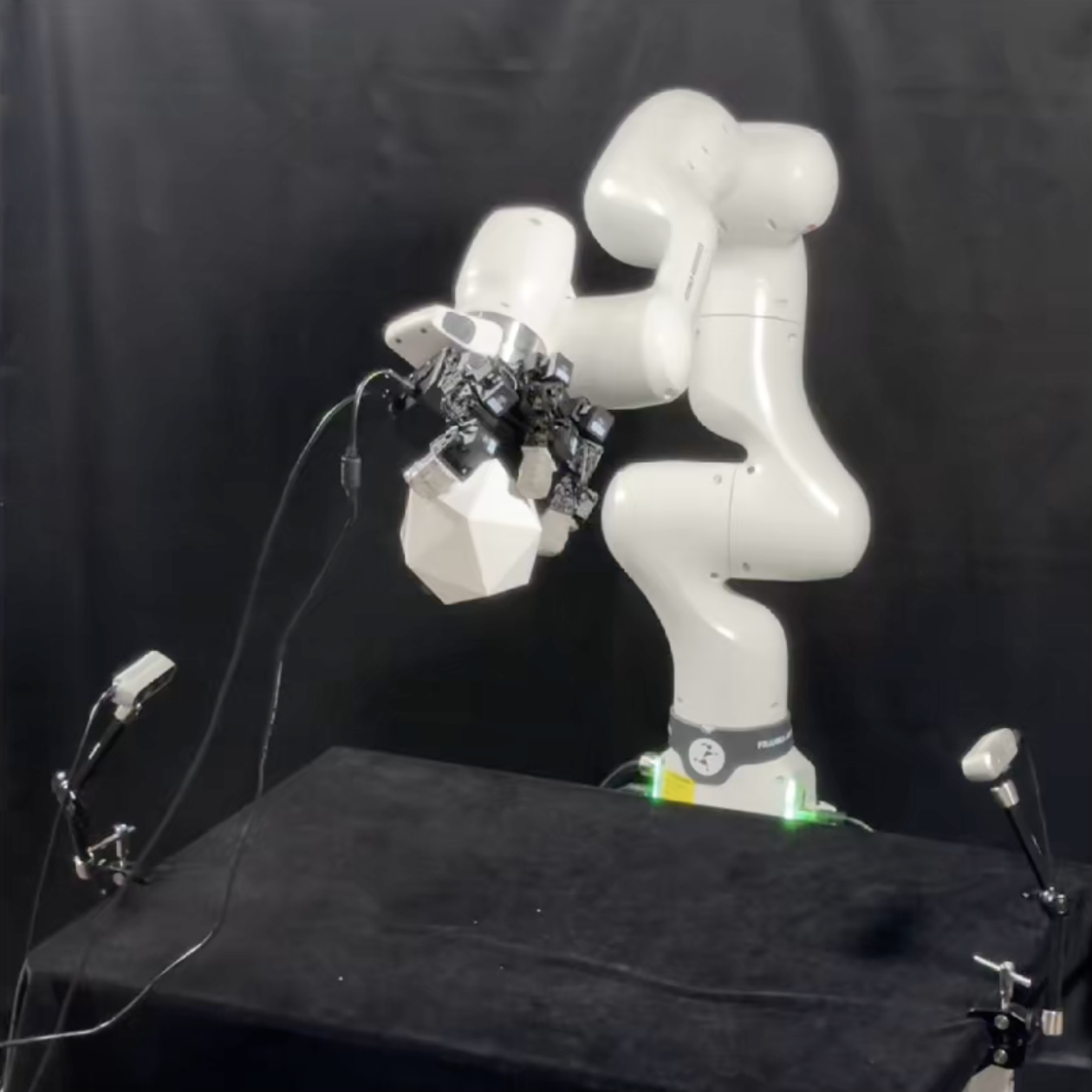}
    \label{fig:icosahedron}
    }

    \subfigure[Grasping a smaller cat toy]{
        \includegraphics[width=0.15\linewidth]{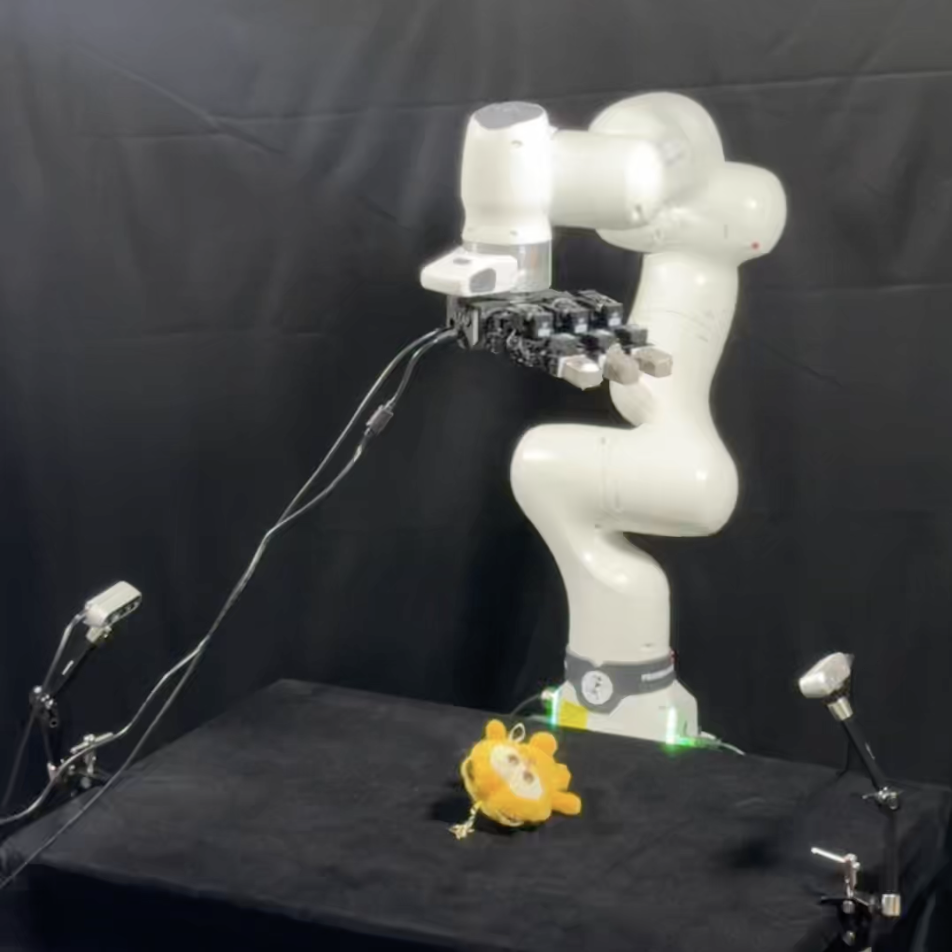}
    \includegraphics[width=0.15\linewidth]{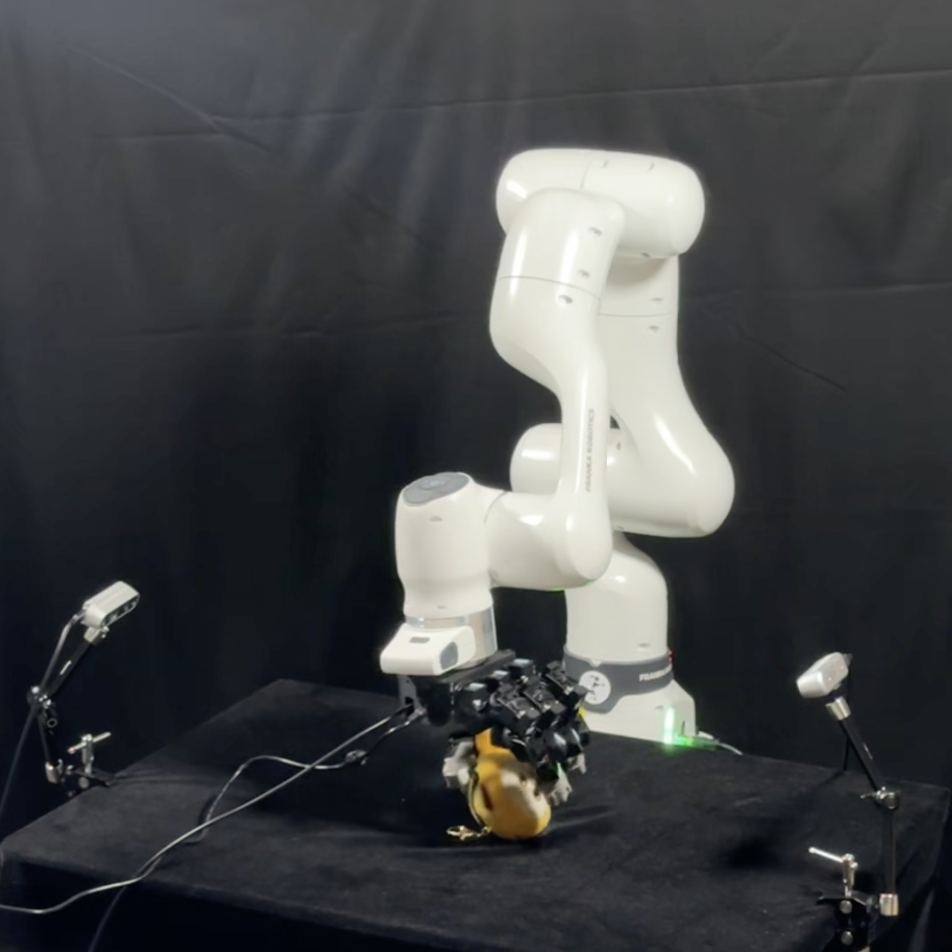}
    \includegraphics[width=0.15\linewidth]{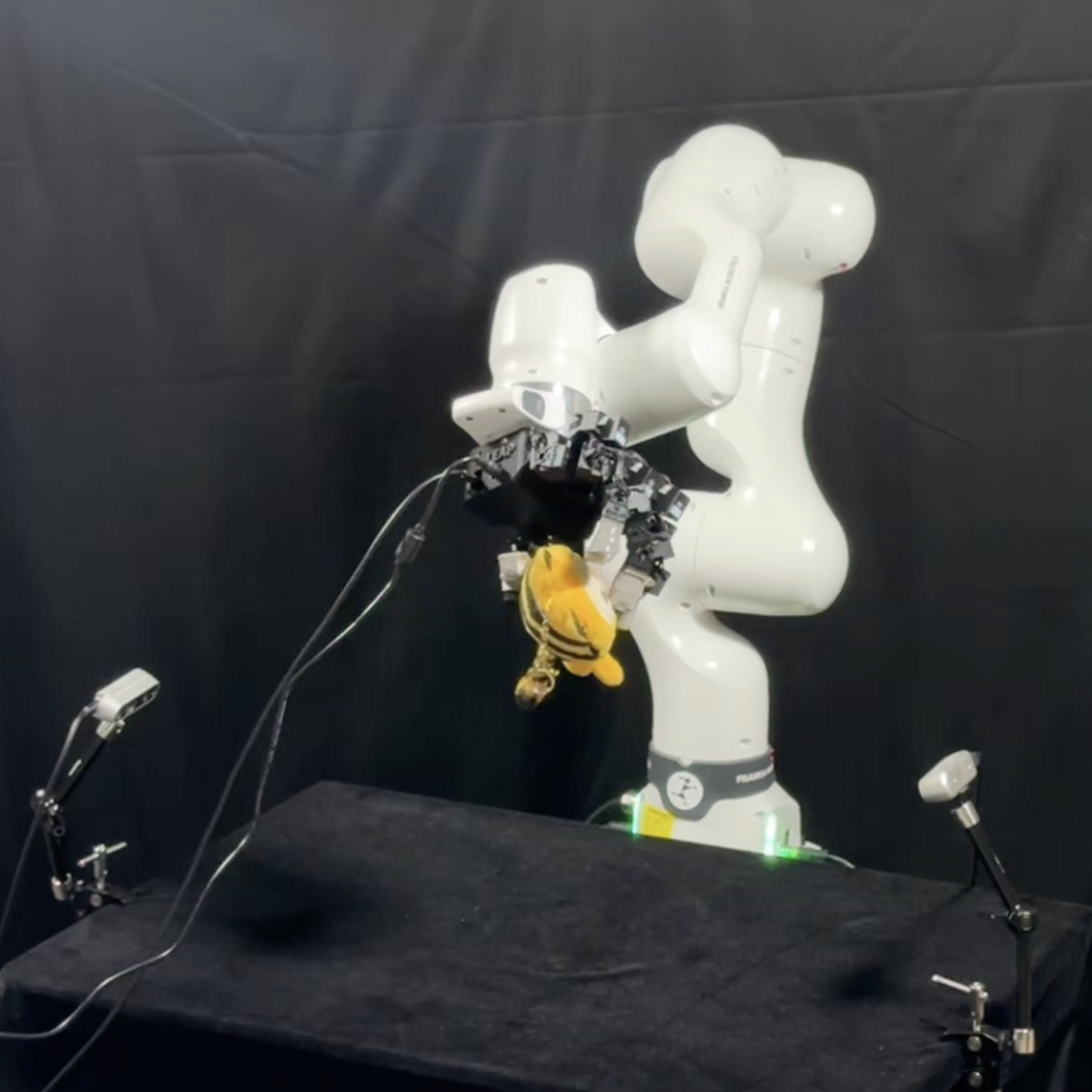}
    \label{fig:cat}
    }
    \subfigure[Grasping a package box in the middle]{
        \includegraphics[width=0.15\linewidth]{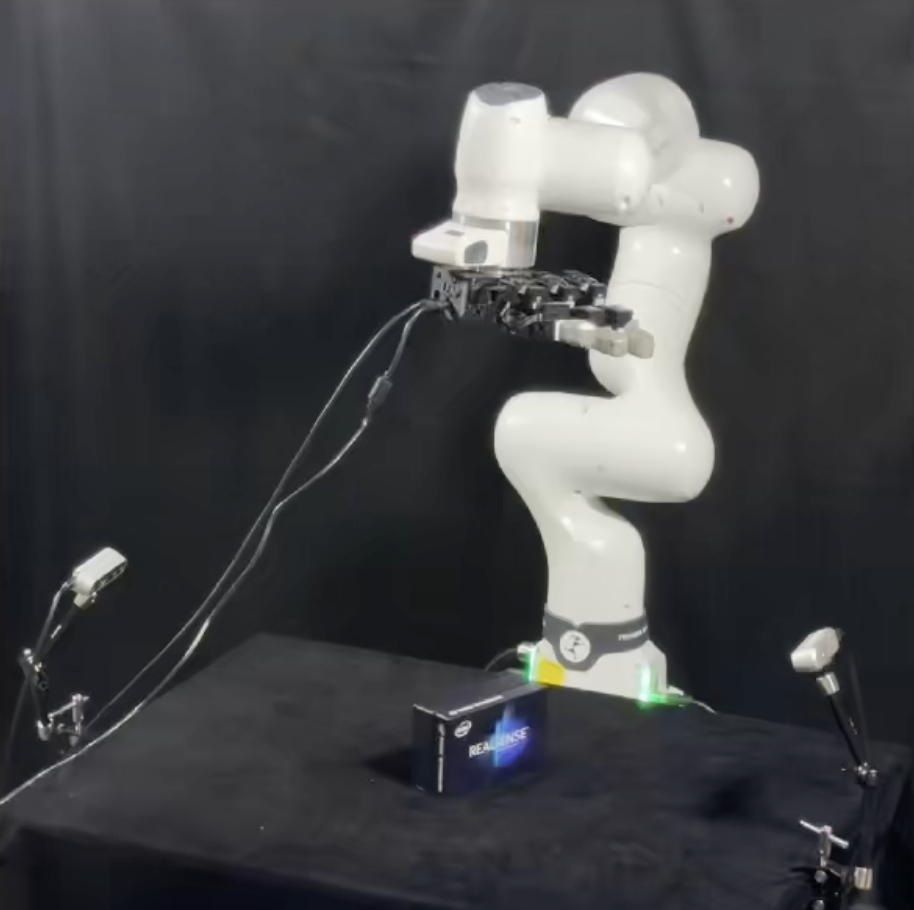}
    \includegraphics[width=0.15\linewidth]{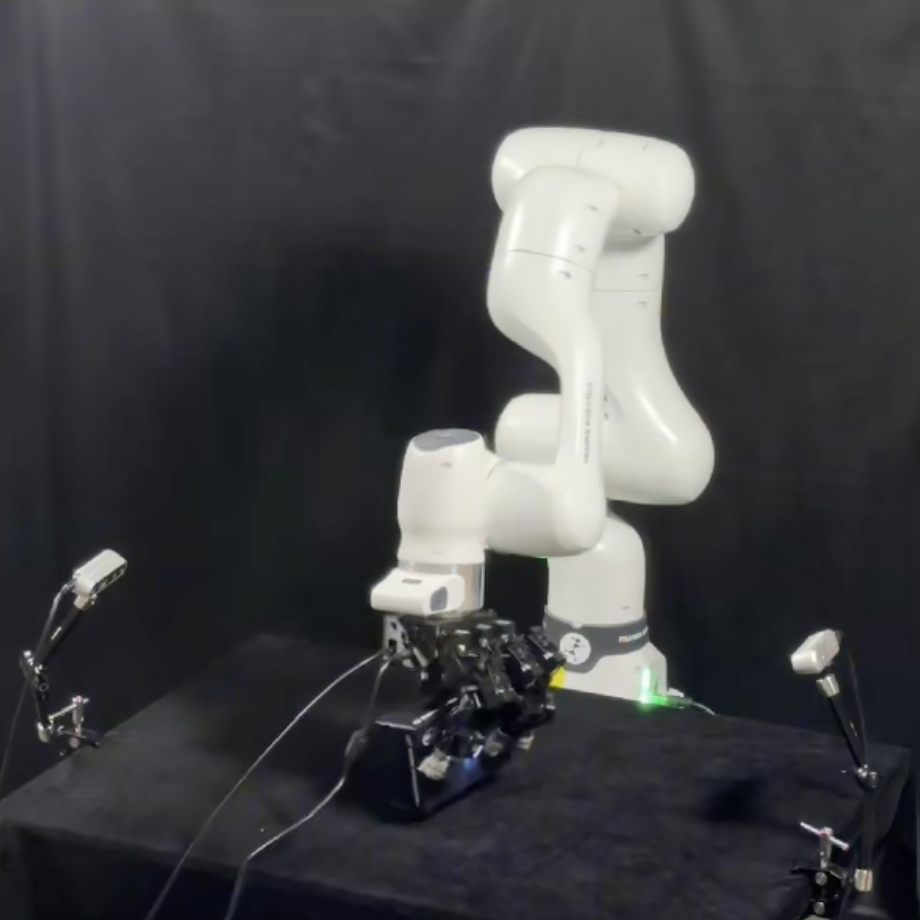}
    \includegraphics[width=0.15\linewidth]{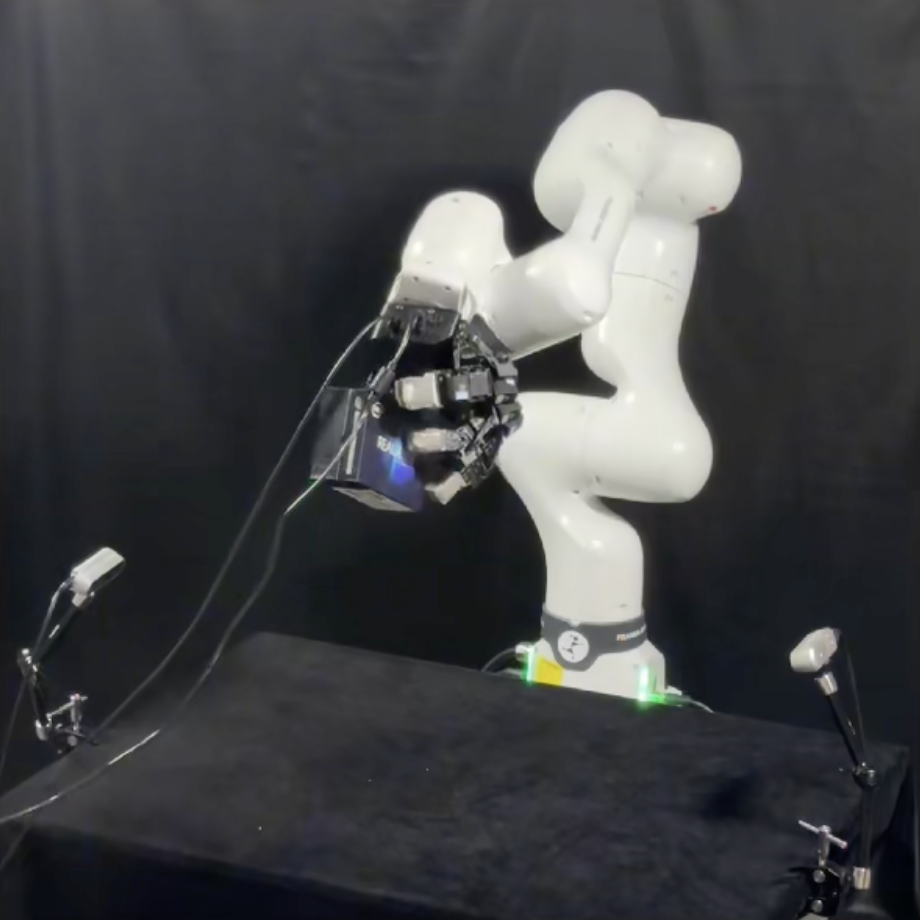}
    \label{fig:box_mid}
    }
    
    \subfigure[Grasping a package box to the left]{
        \includegraphics[width=0.15\linewidth]{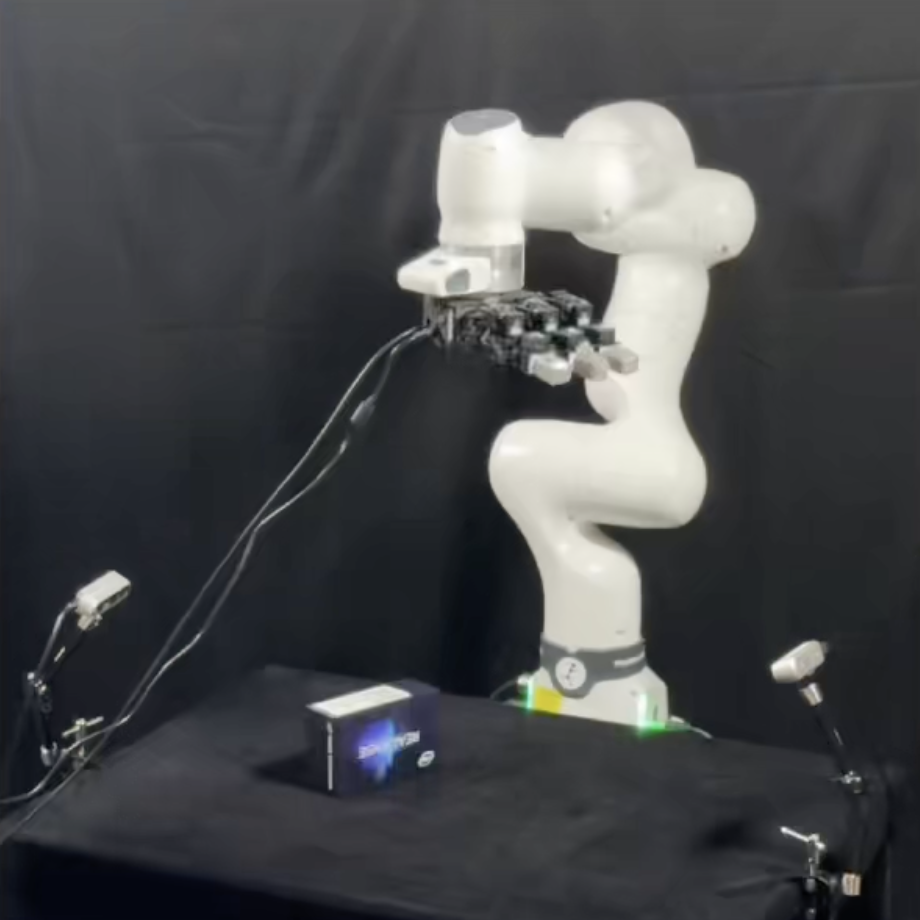}
    \includegraphics[width=0.15\linewidth]{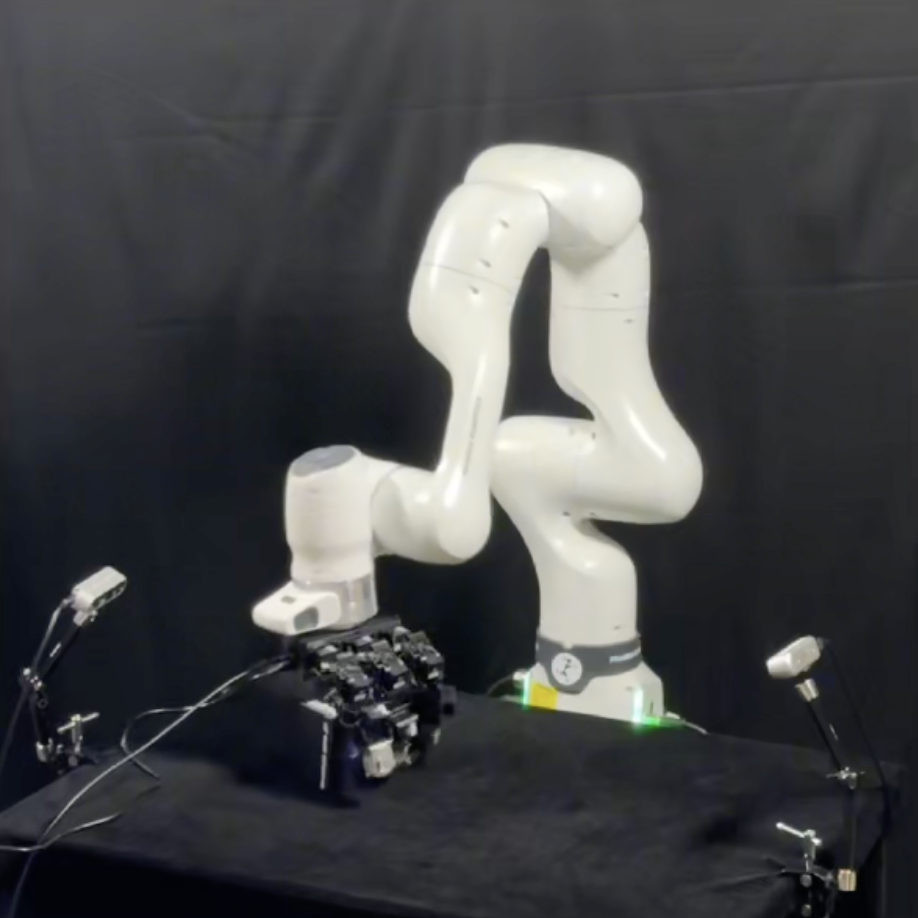}
    \includegraphics[width=0.15\linewidth]{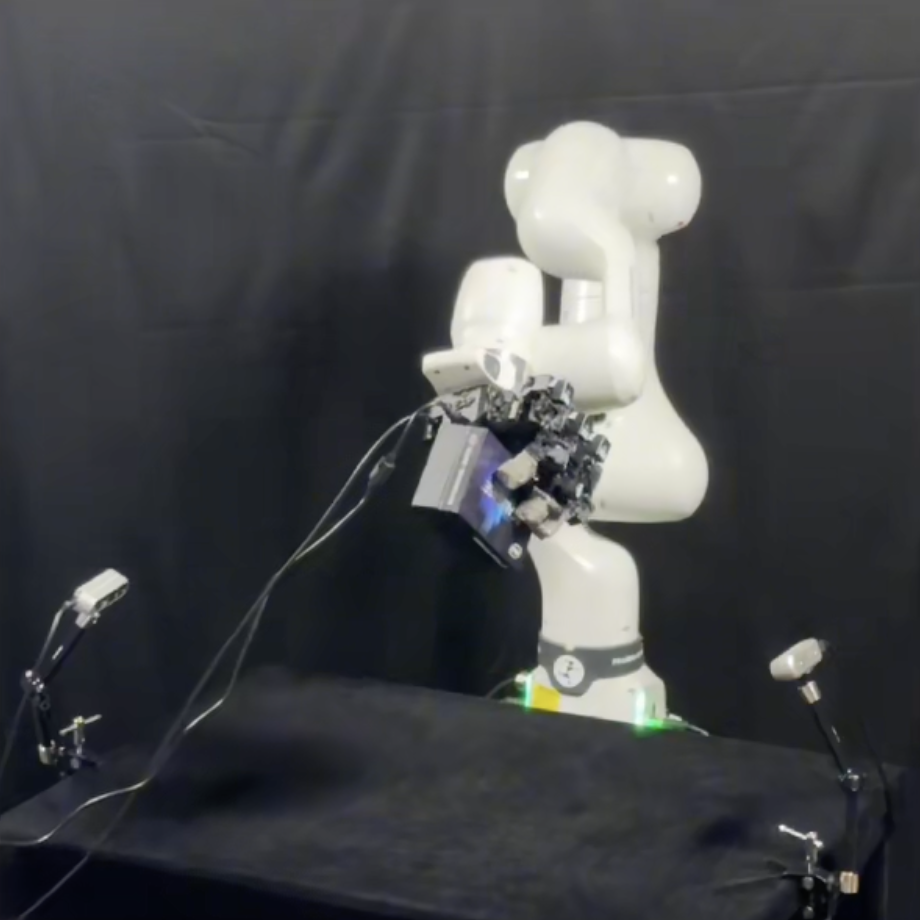}
    \label{fig:box_left}
    }
    \subfigure[Grasping a package box to the right]{
        \includegraphics[width=0.15\linewidth]{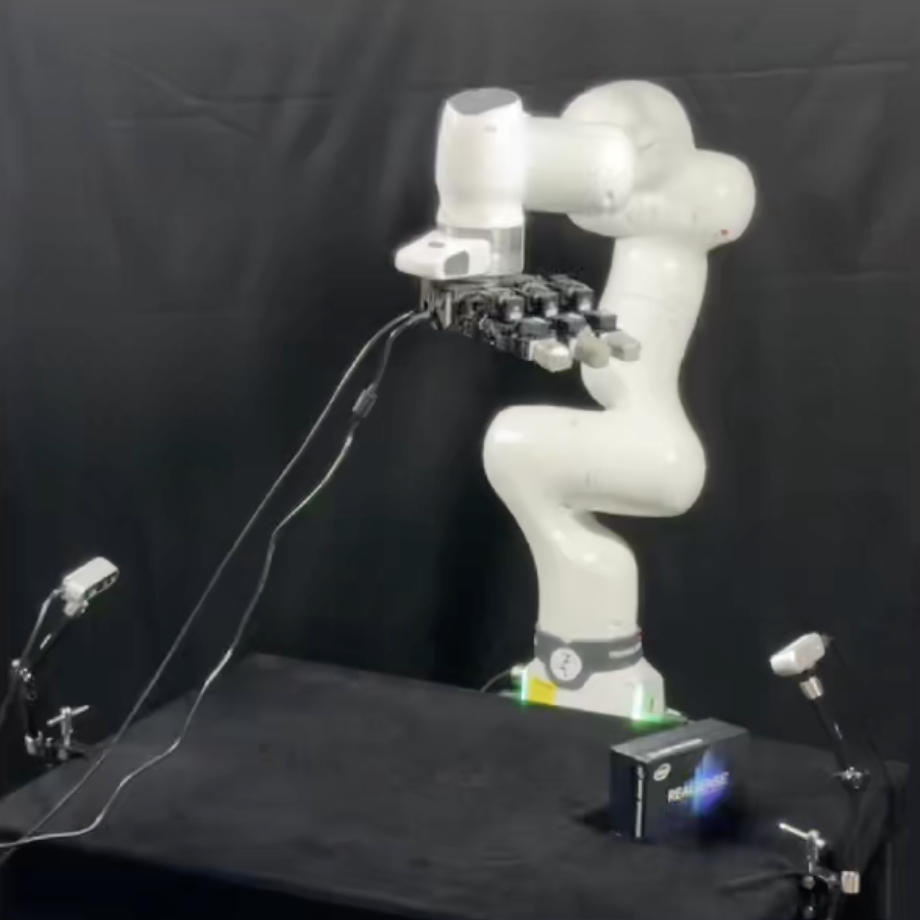}
    \includegraphics[width=0.15\linewidth]{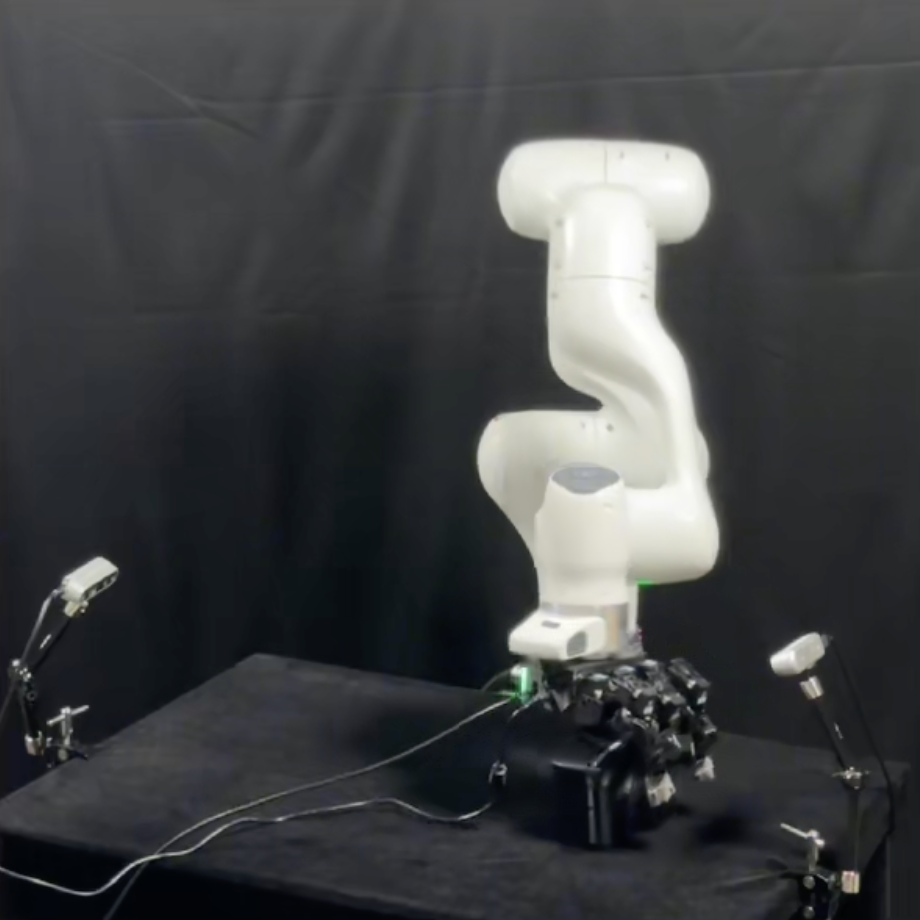}
    \includegraphics[width=0.15\linewidth]{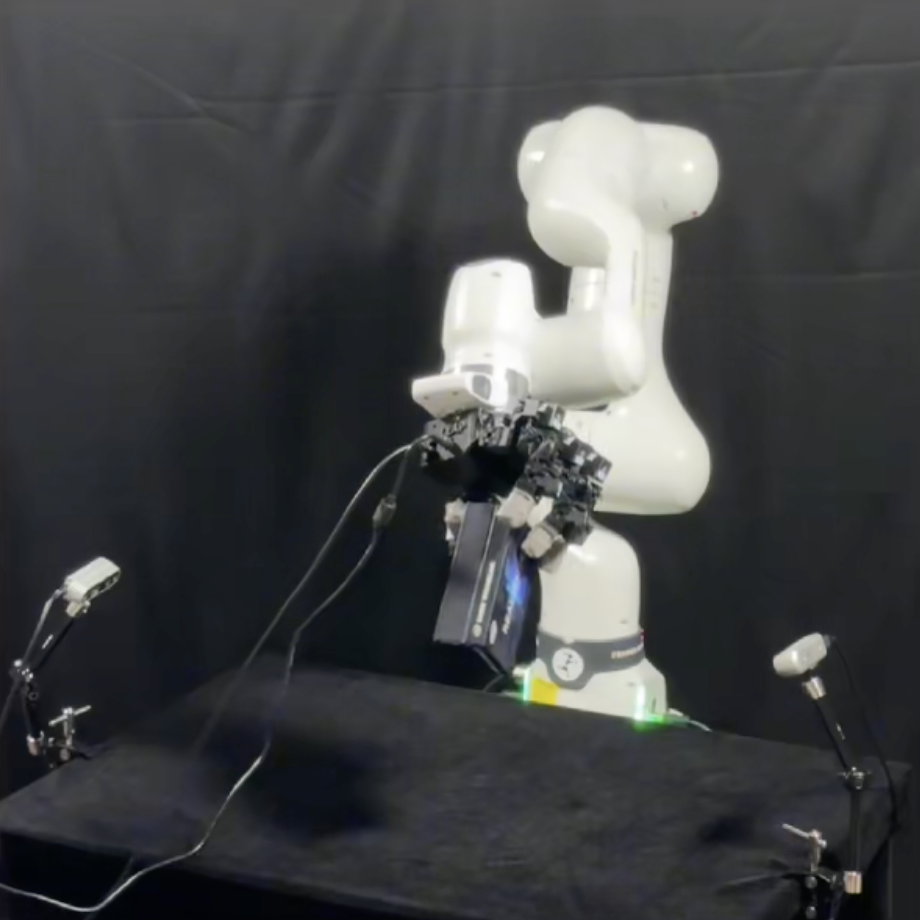}
    \label{fig:box_right}
    }
    \caption{Qualitative grasping results on canonical LEAP hand grasping different objects.}
    \label{fig:qualitative}
\end{figure*}

We first test our distilled DexFormer Policy on canonical LEAP hand. The qualitative grasping results are shwon as Fig.~\ref{fig:qualitative}. The real-world quantitative results are listed in TABLE~\ref{tab:quantitative_success}, showing the success rate and execution average time.

\begin{table}[ht]
\centering
\small
\setlength{\tabcolsep}{6pt}
\caption{Grasping success rate across objects.}
\label{tab:quantitative_success}
\begin{tabular}{lcc}
\toprule
\textbf{Object} & \textbf{Success Rate} & \textbf{Avg. Time (sec)} \\
\midrule
Orange fruit     & $4 / 5$ & 7.45 \\
Red mug     & $2 / 5$ & 6.32 \\
Yellow cube     & $3/5$ & 7.34 \\
Dodecahedron   & $1/5$ & 6.10 \\
Icosahedron    & $2/5$ & 6.45 \\
Large hamster     & $2/5$ & 4.93 \\
Cat toy     & $3/5$ & 6.23 \\
Package box     & $3/5$ & 7.90 \\
\bottomrule
\end{tabular}
\end{table}

\begin{figure*}[h!]
\centering   
\includegraphics[width=0.8\linewidth]{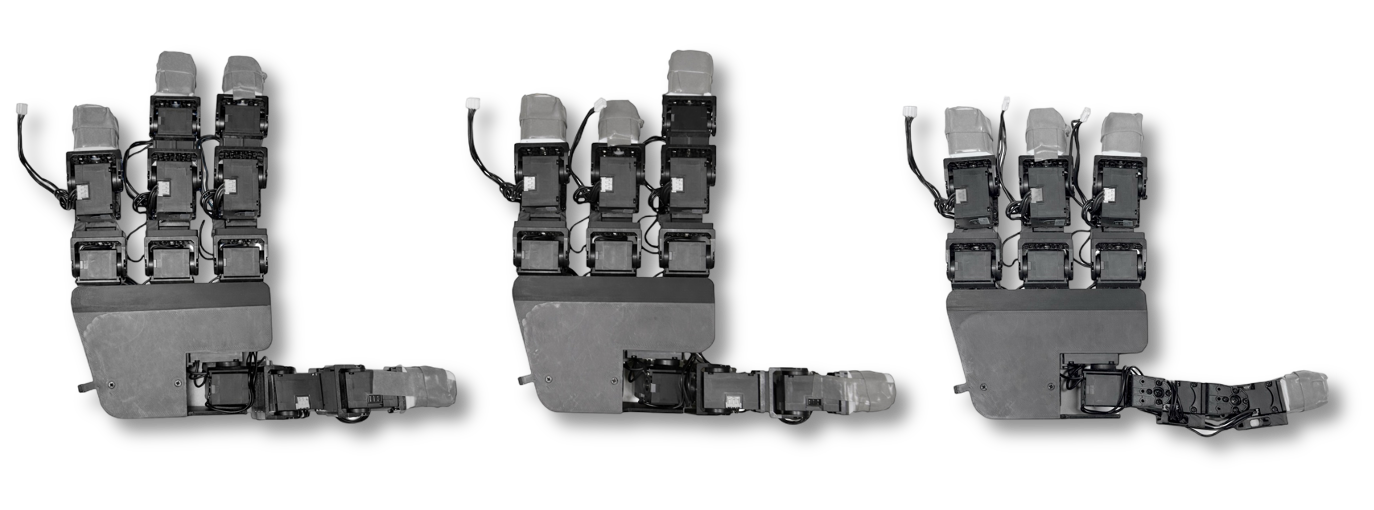}
\caption{Real-world LEAP hand variants with reduced degrees of freedom.}
\label{fig:reduced_viz}
\end{figure*}

\begin{figure*}[h!]
\centering   
\subfigure[LEAP variant with 1 DoF removed.]{
\includegraphics[width=0.25\linewidth]{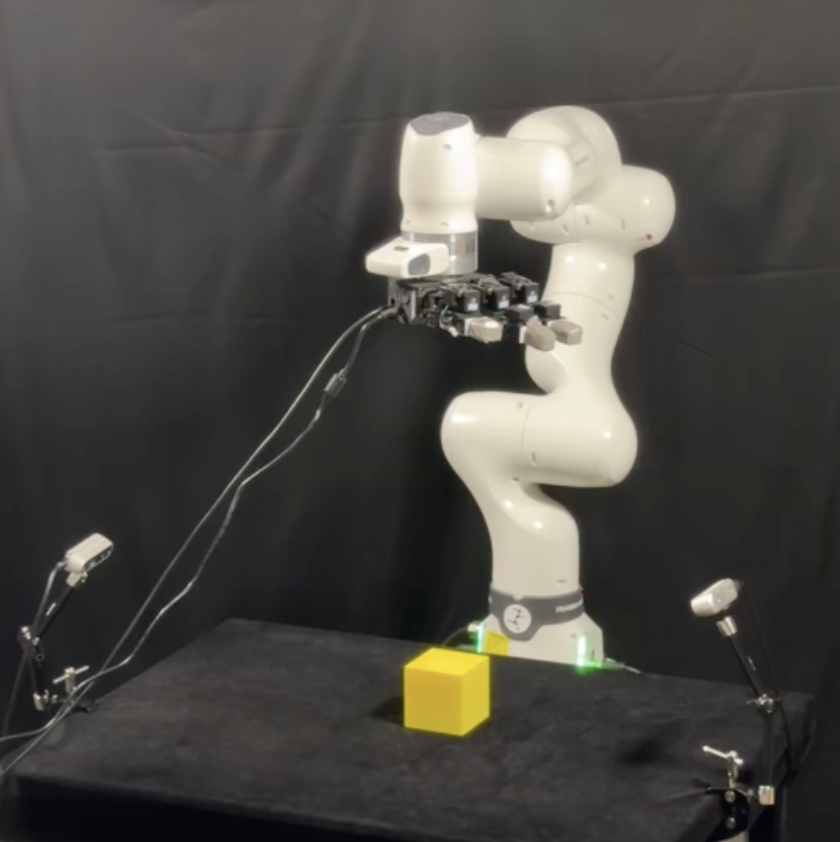}
\includegraphics[width=0.25\linewidth]{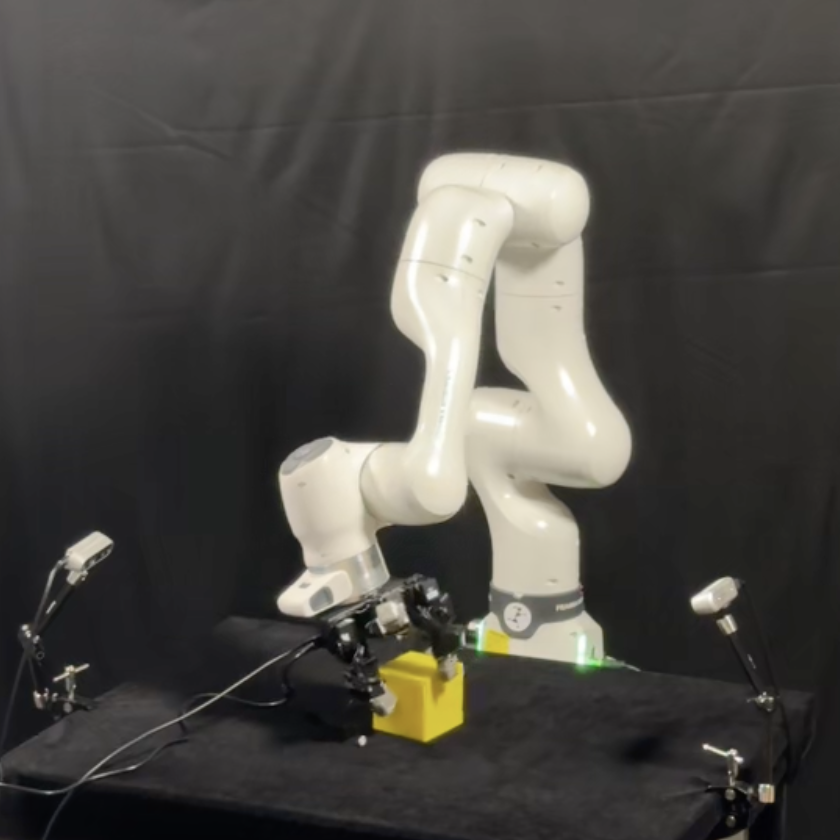}
\includegraphics[width=0.25\linewidth]{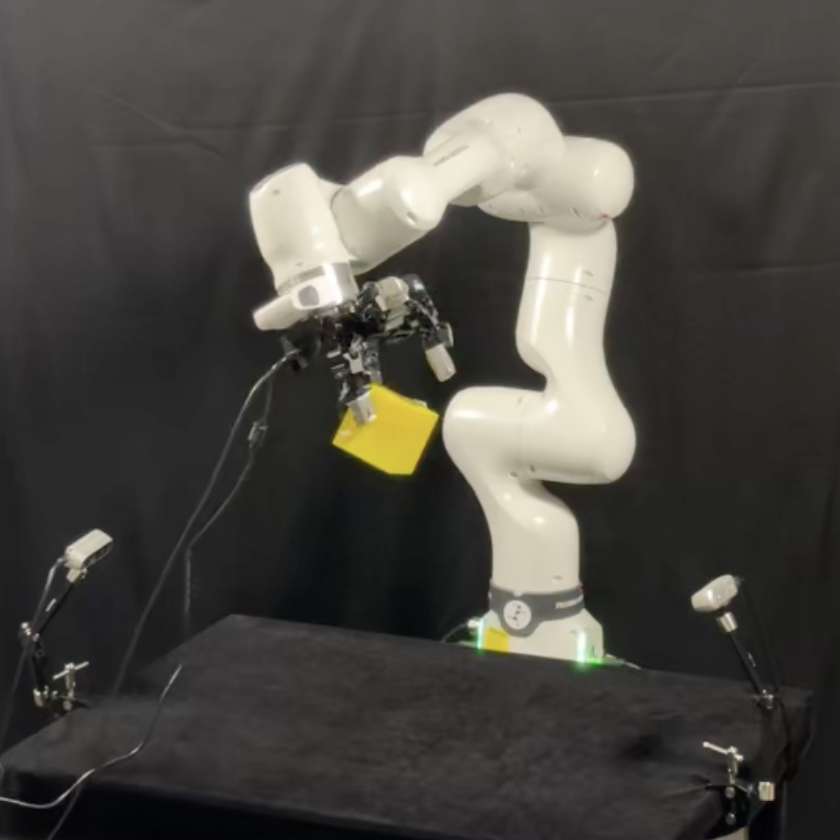}
\label{fig:cut1}
}
\subfigure[LEAP variant with 2 DoFs removed.]{
\includegraphics[width=0.25\linewidth]{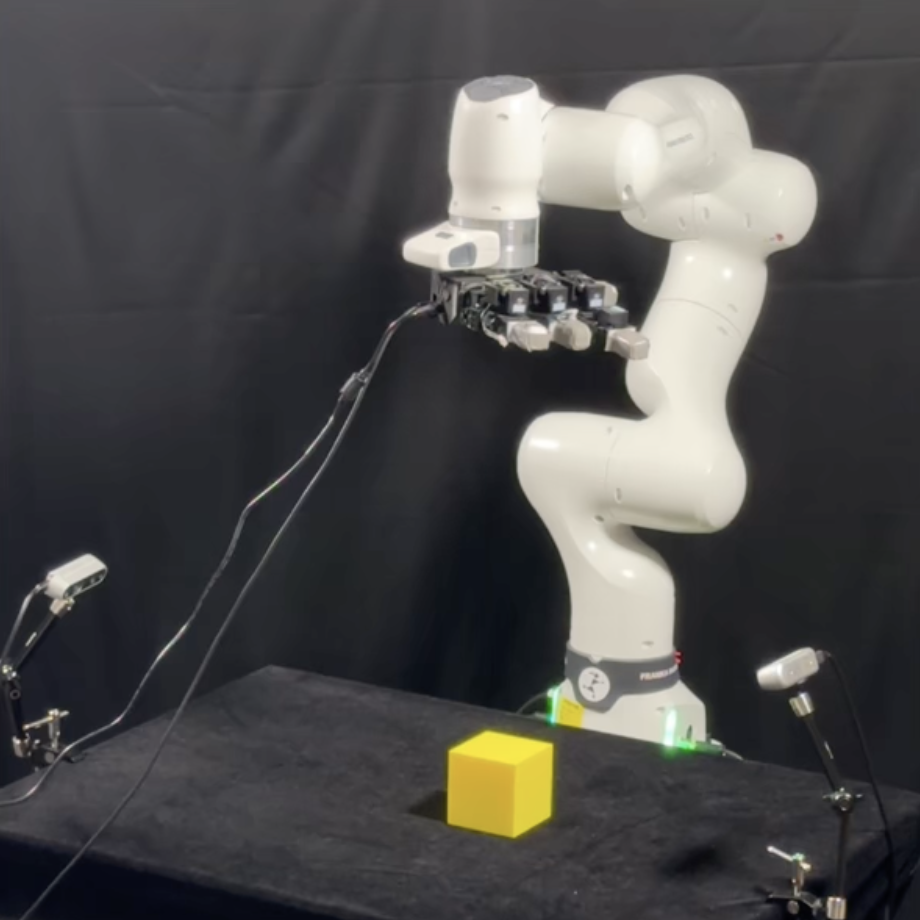}
\includegraphics[width=0.25\linewidth]{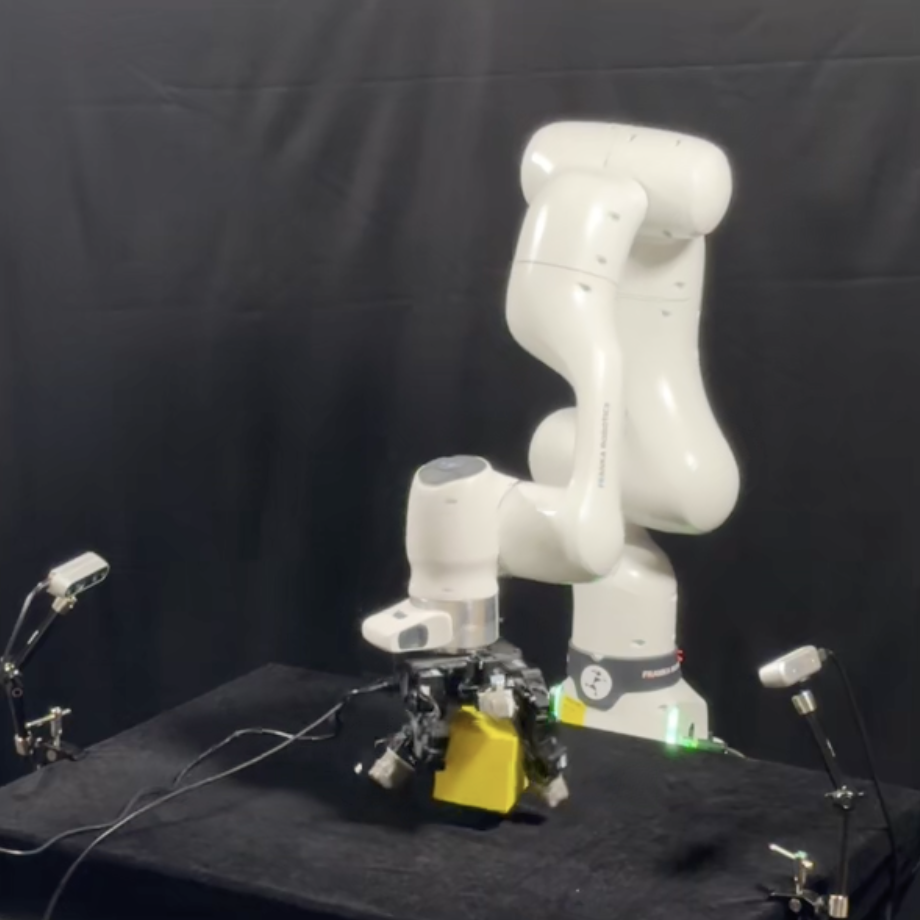}
\includegraphics[width=0.25\linewidth]{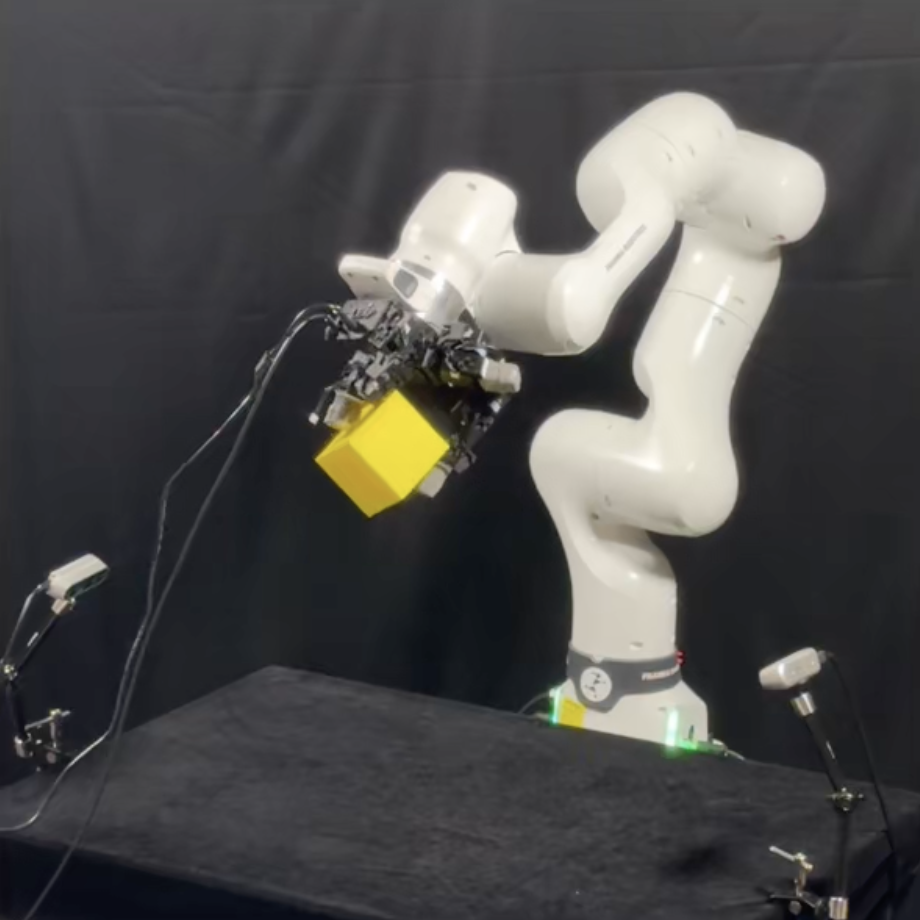}
\label{fig:cut2}
}
\subfigure[LEAP variant with 3 DoFs removed.]{
\includegraphics[width=0.25\linewidth]{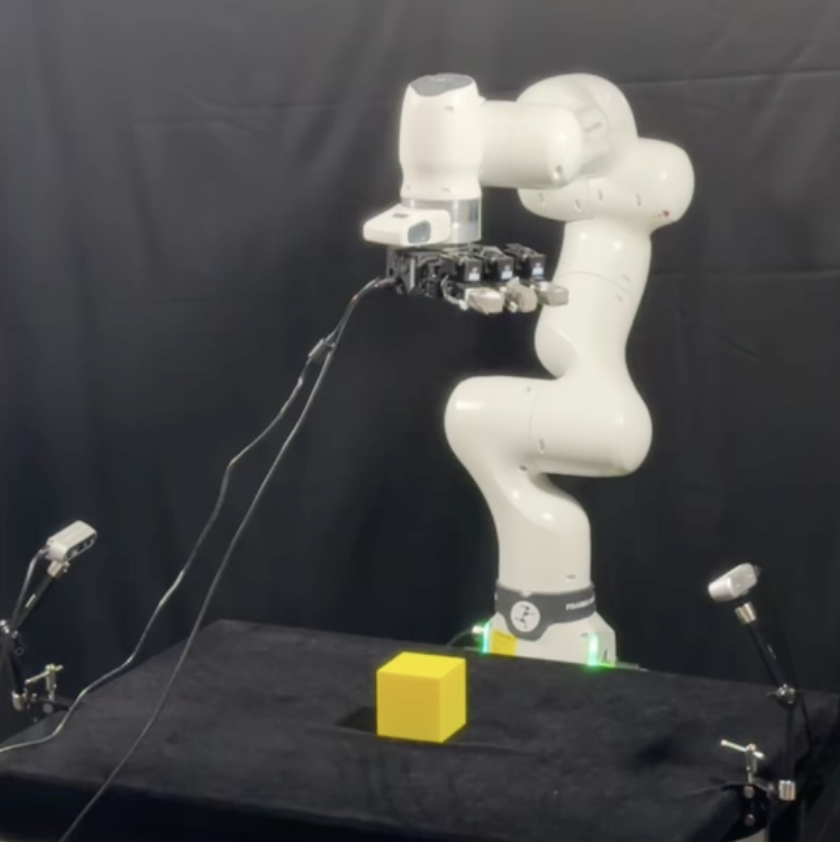}
\includegraphics[width=0.25\linewidth]{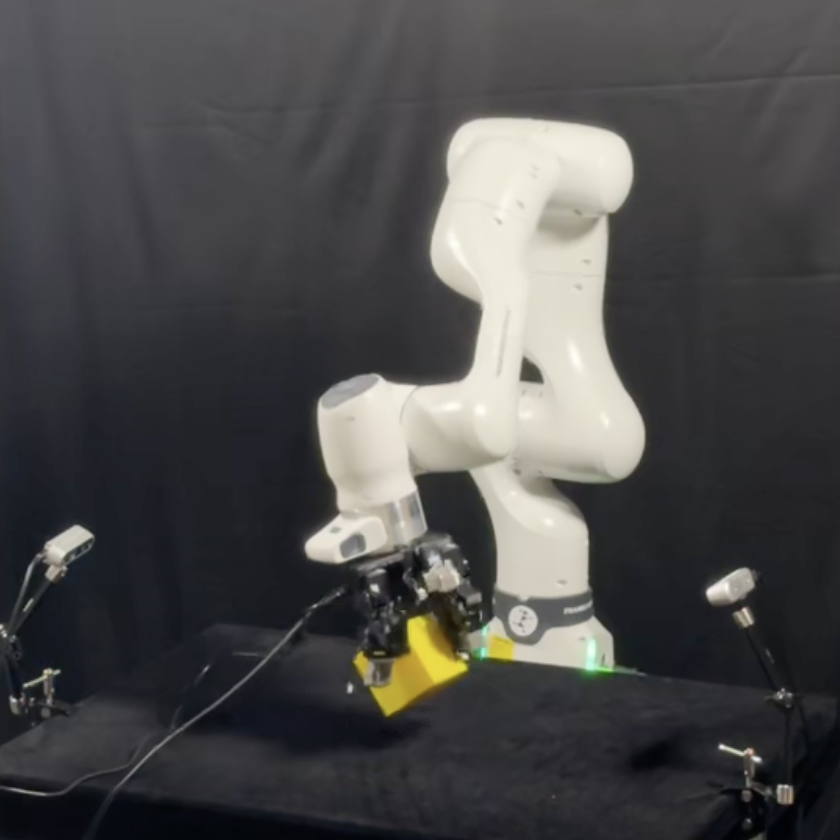}
\includegraphics[width=0.25\linewidth]{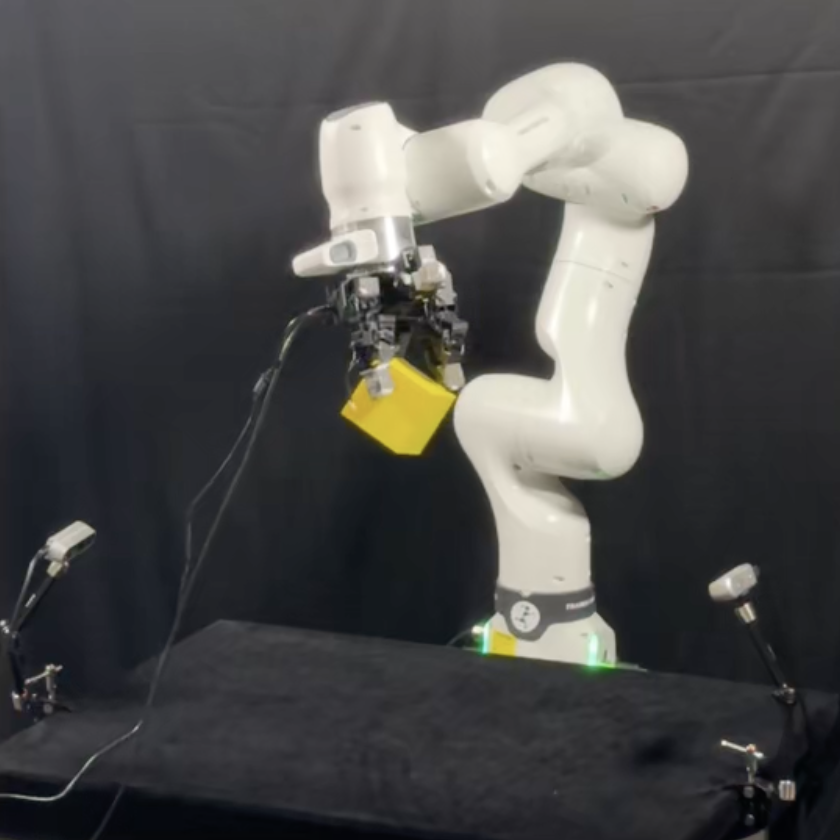}
\label{fig:cut3}
}
\caption{Evaluation on LEAP hand variants with reduced degrees of freedom.}
\label{fig:reduced_viz_eval}
\end{figure*}

We then test a distilled version of DexFormer policy rolled out on three real-world LEAP hand variants. We construct physical LEAP hand variants by removing selected finger joint structures from the canonical configuration, shown in Fig.~\ref{fig:reduced_viz}. From left to right, the variants remove 1 DoF on the ring finger, 2 DoFs on the ring and middle fingers, and 3 DoFs on the index, middle, and ring fingers, respectively. The qualitative grasping results are shown in Fig.~\ref{fig:reduced_viz_eval}.

\end{document}